# Recommending Multiple Criteria Decision Analysis Methods with A New Taxonomy-based Decision Support System

Marco Cinelli[1,*], Miłosz Kadziński[1], Grzegorz Miebs[1], Michael Gonzalez[2], Roman Słowiński[1,3]

[1] *Institute of Computing Science, Poznan University of Technology, Piotrowo 2, 60-965 Poznań, Poland (marco.cinelli@put.poznan.pl; milosz.kadzinski@cs.put.poznan.pl; gregorz.miebs@cs.put.poznan.pl; roman.slowinski@cs.put.poznan.pl)*

[2] *Environmental Decision Analytics Branch, Land Remediation and Technology Division, Center for Environmental Solutions and Emergency Response, Office of Research and Development, U.S. Environmental Protection Agency, 26 West Martin Luther King Dr., Cincinnati, 45268, Ohio, USA (gonzalez.michael@epa.gov)*

[3] *Systems Research Institute, Polish Academy of Sciences, Newelska 6, 01-447 Warsaw, Poland*

[*] *Corresponding author: marco.cinelli@put.poznan.pl*

## Abstract

We present the Multiple Criteria Decision Analysis Methods Selection Software (MCDA-MSS). This decision support system helps analysts answering a recurring question in decision science: "*Which is the most suitable Multiple Criteria Decision Analysis method (or a subset of MCDA methods) that should be used for a given Decision-Making Problem (DMP)?*". The MCDA-MSS includes guidance to lead decision-making processes and choose among an extensive collection (>200) of MCDA methods. These are assessed according to an original comprehensive set of problem characteristics. The accounted features concern problem formulation, preference elicitation and types of preference information, desired features of a preference model, and construction of the decision recommendation. The applicability of the MCDA-MSS has been tested on several case studies. The MCDA-MSS includes the capabilities of (i) covering from very simple to very complex DMPs, (ii) offering recommendations for DMPs that do not match any method from the collection, (iii) helping analysts prioritize efforts for reducing gaps in the description of the DMPs, and (iv) unveiling methodological mistakes that occur in the selection of the methods. A community-wide initiative involving experts in MCDA methodology, analysts using these methods, and decision-makers receiving decision recommendations will contribute to expansion of the MCDA-MSS.

## 1. Introduction

This paper proposes a Decision Support System (DSS) for selecting a Multiple Criteria Decision Analysis (MCDA) method, or a subset of these methods, relevant for a particular Decision-Making Problem (DMP). MCDA is a scientific process aiming to frame DMPs and develop a comprehensive assessment of alternatives (Roy 1990, Tsoukiàs 2007, Cinelli 2017). It has two key potentials. First, it allows identifying, with structured and traceable protocols, the alternatives to be considered and the criteria to evaluate them. Second, it enables conveying a wealth of information that describes each alternative in a synthetic fashion, like a ranking from the best to the worst, a sorting of good, medium, and bad classes, or the choice of a subset of the most preferred alternatives.

There are several complexities involved in conducting the MCDA process. These include (i) the framing of the decision situation to be studied (Ley-Borrás 2015), (ii) generation and characterization of alternatives to be considered (Keeney 1996), (iii) development and identification of evaluation criteria (Keeney and Gregory 2005), and (iv) selection of MCDA methods for each case study (Wątróbski et al. 2019). Due to these complexities, tools are needed to aid the MCDA-based research. This contribution is primarily focused on the latter complexity, and the presented DSS aims to support analysts in the choice of the MCDA method(s) for each case study.

Over the last few decades, the number of MCDA methods has grown steadily (Hwang and Yoon 1981, Wallenius et al. 2008, Greco et al. 2016a, Alinezhad and Khalili 2019), and an analyst can find it challenging to select the most suitable method. The main issue that a decision analyst is faced with is summarized by this question:

“*Which is the most appropriate MCDA method (or a subset of methods) that should be used for a given DMP?”*

This dilemma and one possible solution are presented in Figure 1. The analyst is usually in a situation where the characteristics of the DMP must be accounted for in order to lead to selecting the MCDA method(s) that fit best with the DMP.

***Figure 1:*** *The challenge faced by decision analysts when the DMP is structured*

These characteristics are the features or elements that define the MCDA process. As seen in Figure 1, the elements (light blue boxes) are: (i) desired recommendation is a ranking of the alternatives, (ii) criteria are structured hierarchically, (iii) the evaluation scales of the criteria are deterministic, and (iv) weights of the criteria are exact trade-offs. Thus, the challenge consists in finding the MCDA method(s) that can support these characteristics, which with this simple example could be a weighted sum with normalization (Itsubo 2015) or MAVT (von Winterfeldt and Edwards 1986). In other words, the challenge is to obtain a match between the DMP characteristics and the MCDA method(s) that can address such characteristics or at least satisfy as many of them as possible.

To avoid overwhelming the analysts and Decision Makers (DMs) with a wide pool of methods and/or (ii) prevent wrongly choosing the method(s) (or at least not selecting the most suitable one), tools are needed to assist analysts in choosing an MCDA method(s) for each specific application. It is essential to also add that the consequences of choosing an inadequate method among the plethora of those available are substantial. This can lead to neglecting some critical aspects of the problem, undesired compromises, and ultimately lead to a recommendation not aligned with the actual problem's characteristics and preferences of the involved stakeholders (Ebert and Welsch 2004). An additional warning sign is the fact that popularity, simplicity, and intuitiveness are among the main reasons for selecting MCDA methods (Cajot et al. 2017). These are some key reasons that drove the emergence of DSSs to aid in selecting MCDA method(s), which are briefly reviewed in the next section.

## 1.1. Review of DSSs for MCDA method(s) recommendation

MCDA methods selection is an MCDA problem in its own right (Gershon 1981, Guarini et al. 2018). It involves the MCDA methods as the alternatives and their decision support capabilities as the evaluation criteria. The most up-to-date comprehensive review of the available DSSs for MCDA method(s) recommendation has been recently presented by Cinelli et al. (2020a). They analyzed 23 of these DSSs and clustered them into three groups: (i) rules-based, (ii) algorithm-based, and (iii) artificial neural network-based. It was found that each DSS is structured on a formal representation of the methods, which can be called a taxonomy. In these DSSs, the taxonomy contains a set of characteristics (i.e., features) that describe the MCDA methods and the types of decision-making challenges they can support solving. For example, Celik and Topcu (2009) investigated the type of supported problem statement, measurement scale, weights, and thresholds. Benoit and Rousseaux (2003) emphasized the effect of the level of compensation in each method and the sensitivity to thresholds. Gershon and Duckstein (1983) highlighted the qualitative features, like the easiness of use, processing time needed to compile the data required for the method, and alternatives and/or criteria the MCDA method can work with. Salinesi and Kornyshova (2006) focused on the possible dynamic character of the DMP, the structure of the family of criteria (i.e., flat or hierarchical), and the type of preference models.

### 1.2. Motivation: What's missing in the existing DSSs for MCDA method(s) recommendation?

DSSs for MCDA method(s) recommendation require their own criteria to lead the selection process. As indicated by Cajot et al. (2017), a system capable of describing the MCDA methods and the practical implications of using one method rather than another is pivotal for these DSSs. This requires the systematic axiomatization of MCDA methods, which was an issue that already emerged in the 1990s' (French 1993), but has only been addressed partially thus far. An initial contribution to achieving this ambitious target was presented in the recent review by Cinelli et al. (2020a), which analyzed 56 peer-reviewed publications that consider the features (called characteristics) that should be accounted for when conducting an MCDA process and leading to the selection of an MCDA method or a subset of these methods. This work resulted in a comprehensive taxonomy to describe the MCDA process as well as its methods. The taxonomy is composed of 10 main characteristics (and sub-characteristics) clustered into three phases. The first is the problem formulation phase, which examines problem typology, the structure of the criteria, and evaluating the performance of alternatives. The second phase is focused on how the decision recommendation is developed with different strategies to elicit preferences of the DMs, among which the type of weights, thresholds, aggregation functions, and indirect elicitation approaches. The third phase is focused on the technical support to handle the problem with an MCDA software and the qualitative features of the DMP describing how complex and flexible the method is, its data preparation requirements, and its reported use in the relevant literature.

The available DSSs include a limited set of taxonomy characteristics to describe the MCDA methods and their decision support capabilities (Cinelli et al. 2020a). This means that they do not account for several key problem characteristics that analysts regularly deal with, making it difficult to identify the relevant MCDA method(s) in real case studies. These include, among others, the presence of hierarchies in the set of criteria, the type of preferences provided by the decision makers (i.e., direct or indirect), a wide array of uncertainty analyses, the mathematical foundations of the underlying algorithms, and the capacity to handle inconsistent and/or dynamic datasets and preferences. Consequently, the type of MCDA problems that can be tackled and the way MCDA methods can be described with the available DSSs are limited to a subset of the proposed features in Cinelli et al. (2020a). In addition, the available DSSs consider a rather limited group of methods, ranging from five (Li et al. 2008, Eldrandaly et al. 2009) to 56 (Wątróbski et al. 2019). This means a large share of the most recent and advanced methods is not included in the available DSSs. What is more, the available DSSs do not face critical issues that can emerge when searching for an MCDA method for a particular case study. These include the cases when (i) no method is recommended or (ii) many methods are recommended. In the first case, the analyst finds himself in a situation where no decision support is provided at all, while in the second one, too general advice is given.

### 1.3. Main contributions of the paper

This paper proposes a DSS called Multiple Criteria Decision Analysis Methods Selection Software (MCDA-MSS) that recommends the most suitable MCDA method(s) for a given DMP. This DSS has two aims: (i) to allow describing complex DMPs and distinguish many methods proposed for MCDA by means of a set of relevant features (characteristics), and (ii) to guide an analyst assisting a DM in choosing the most appropriate method(s) for a given MCDA problem. The MCDA-MSS is now available for free at the following link http://mcdamss.com. The type of DMP we consider involves one-stage decision by a single decision maker, concerning a finite set of alternatives with known consequences, evaluated by a finite set of conflicting criteria.

The users of this MCDA-MSS are envisioned to be analysts who understand how complex decision-making based on multiple criteria can be conducted and the methods that can aid it. However, they acknowledge that human cognition alone is not enough to deal with the complexities of current decision-making challenges. For this reason, they can make use of the structured decision support methodology that is embedded in the MCDA-MSS to better inform their selection of MCDA methods.

Compared to the existing DSSs, the MCDA-MSS provides several unique contributions. First, it uses a comprehensive taxonomy of 156 characteristics to describe MCDA methods within its library, updating the taxonomy presented in Cinelli et al. (2020a), which included "only" 66. The taxonomy in the MCDA-MSS thus represents the most detailed vocabulary that can be used to describe and develop MCDA methods and guide complex DMPs. Second, it includes a large set of MCDA methods, more than 200, which represent different approaches, schools, tendencies, and methodological streams that have evolved in the field of MCDA over the past sixty last. As a result, we offer the most comprehensive database of MCDA methods, with their number being almost four times greater than in the state-of-the-art decision support system elaborated by Wątróbski et al. in 2019. Third, it provides solutions to cases where no method matches all the requirements set by the analysts. These solutions involve a dialogue with the DM concerning demands that must be fulfilled by the recommended methods and aiming at maximizing the share of requirements that can be satisfied. Fourth, it offers a strategy to reduce a large set of relevant methods when there is much uncertainty in the description of the DMP. It suggests the questions that maximize the potential information gain from the users' answers by minimizing the number of recommended methods once a desired feature is specified. Fifth, it can be used as an identifier of errors in MCDA method selection. This capacity is illustrated by studying ten case studies reported in the literature and discussing the reasons for mismatches of MCDA methods.

## 2. MCDA-MSS development

The methodology formulated to develop and test the MCDA-MSS included three stages (see Figure 2). The first stage concentrates on shaping the taxonomy used within the MCDA-MSS to describe each MCDA method in its library. The second stage is tailored to the development of the MCDA-MSS in a web-software form, while the last stage focuses on testing the MCDA-MSS on a series of case studies to assess its usability and performance. Each stage is described in detail in the following sections.

***Figure 2:** The methodology used to develop and test the MCDA-MSS*

## 2.1 Stage 1: Develop the taxonomy and the database of MCDA methods used in the MCDA-MSS

Stage 1 was focused on (i) the development of the methodological backbone of the MCDA-MSS, i.e., the taxonomy of characteristics (i.e., features) used to describe the MCDA methods, and (ii) the database of MCDA methods themselves. This was achieved in two steps. The first one was the study of the taxonomy introduced in Cinelli et al. (2020a), called here *taxonomy v.1*. The second one consisted in its application to a large set of MCDA methods. The rationale for the selected methods, together with their brief description, is given in Section 2.1.2 and Appendix A in the Electronic Supplementary Information (ESI). An important remark on the database of methods in the MCDA-MSS is the objective was not to include all the available MCDA methods, but rather propose an initially wide list, which can be complemented in the future. Also, the current focus has been on methods for a single DM, while those for group DMs can be a topic of further inclusion (see Section 4.2 for future research avenues).

### *2.1.1 The taxonomy used in the MCDA-MSS*

The application of taxonomy v.1 from Cinelli et al. (2020a) to the MCDA methods led to its refinement and revision, resulting in an elaborated form, called *taxonomy v.2,* that has been implemented in the MCDA-MSS. Its structure is shown in Table 1, introducing multiple novelties when compared to the one of Cinelli et al. (2020a). This one expanded and also restructured the first two phases, while the third one was not included. The reason for such exclusion is that the MCDA-MSS aims at describing the MCDA methods according to the features that can be evaluated as objectively as possible. Being that the third phase of the taxonomy in Cinelli et al. (2020a) focused on qualitative features, it did not fit with the purpose of the MCDA-MSS. Two examples of these qualitative features are easiness of use of the method and the time required to compile the needed data to use the method.

Their assessment depends on the knowledge and expertise of the analysts who lead the MCDA process, and so they cannot be characterized objectively.

The differences between taxonomy v.1 and v.2 are substantial, starting from the number of considered objective features. Those in taxonomy v.1 of Cinelli et al. (2020a) are 66, while those in taxonomy v.2 of the MCDA-MSS are 156. As a matter of context regarding the most recent software to recommend MCDA methods, the number of features they account for is 9 in Wątróbski et al. (2019) and 39 in Guarini et al. (2018).

The taxonomy in the MCDA-MSS is structured in four main sections:

1. Problem typology: Defines the type and structure of the DMP;
2. Preference model: Defines what type of model the analyst would like to apply;
3. Elicitation of preferences: Defines the type, modality, and frequency of elicited preferences;
4. Exploitation of the preference model: Defines the strategy used to derive and enrich the decision recommendation.

#### *2.1.1.1 MCDA-MSS section 1: Problem typology (c.1)*

In the problem typology section, the analyst can define how the problem is framed by (i) choosing the type of DMP under consideration and (ii) describing the criteria used to assess the alternatives.

As far as the type of decision-making challenge is concerned, the problem statement (c.1.1), in other words, the kind of desired decision recommendation, can be of four types (Belton and Stewart 2002, Cailloux et al. 2007). These include ranking (i.e., order the alternatives from the most to the least preferred), sorting (i.e., assign the alternatives to pre-defined preference-ordered decision classes), clustering (i.e., divide alternatives into groups according to some similarity measure or preference relation), and choice (i.e., select the most preferred subset of alternatives). Except for choice problems, it is also possible to distinguish the type of order of the alternatives/classes/clusters, as either partial or complete (Roy 2016b). Partial ordering admits incomparability, and it does not necessarily lead to a univocal ordering. This does not hold for the complete order, where all the alternatives/classes/clusters are ordered from the most to the least preferred. In addition, the scale leading the recommendation for ranking and sorting problems can also be chosen between ordinal and cardinal (Roy 2016b). Ordinal recommendations are based on binary relations, where only the position of the alternatives is meaningful. In contrast, cardinal recommendations are driven by a score, where the distance between alternatives is meaningful in quantitative terms. As far as sorting and choice are concerned, it is possible to set the cardinality of the DMP, either with or without constraints (Kadziński and Słowiński 2013). Cardinality with constraints consists in the DMP where a pre-defined number of alternatives is either chosen or assigned to each class. On the contrary, cardinality without constraints does not restrict the number of chosen alternatives or assignments to each decision class. The type of set of alternatives (c.1.2) can then also be chosen, being either stable (i.e., no new alternatives are foreseen and added to the set) or incremental (i.e., new alternatives keep arriving as the decision context evolves) (Siebert and Keeney 2015). The assumption that we applied in the development of the MCDA-MSS database is that methods that handle an incremental set of alternatives can also handle one with a stable set. In addition, in case the user chooses an incremental set of alternatives and the problem statements of ranking, choice, sorting with cardinality constraints, and clustering, the MCDA-MSS notes that the recommendation provided by the suitable methods might change when new alternatives are added.

Four features characterize the description of the criteria used to assess the alternatives. The first one is the structure of the set of criteria (c.1.3), being either flat (i.e., the criteria are all at the same level) or hierarchical (i.e., the criteria are organized in levels, hierarchically, from general to detailed ones) (Marttunen et al. 2018). The second is the type of performance of the criteria (c.1.4), which can be either deterministic (i.e., exact input) or uncertain (Keeney and Gregory 2005). A further differentiation here is on whether the performance of an alternative is provided on a criterion per se, or whether it refers to the performance of an alternative on a criterion with respect to the performance of another alternative on the same criterion, like in the Analytical Hierarchy Process (AHP) (Saaty 1980), for example. Uncertain performance includes multiple options, among which the lack of information, intervals, probability distributions, and fuzzy (Dias et al. 2012). The third feature describing the criteria is the preference direction of the criteria (c.1.5), which defines whether the order of preference for the values of the criteria is known or has to be discovered (Błaszczyński et al. 2012). In case it is known, two options are available. The first one is for monotonic criteria (Nardo et al. 2008), whose order of preference is defined as constantly non-decreasing or non-increasing with respect to the performances on the criterion. The second one is for non-monotonic criteria (Pap et al. 2009), whose order of preference is defined as non-decreasing or non-increasing in different regions of the evaluation scale with respect to the performances on the criterion. The taxonomy does not include attributes without preference-ordered scales, as the objective of the MCDA-MSS is to focus on problems that are characterized in classical MCDA terms, where for each criterion, a preference-order can be assumed, even if the positive or negative monotonicity relationships do not hold in the whole evaluation space. The last feature in this section looks at whether the criteria set is complete or not (c.1.6) (Roy 2016b). The set is complete if all criteria relevant for the problem and having an impact on the recommendation were identified at the problem modelling phase and included in the set. In contrast, it is incomplete when at least one criterion relevant for the problem and having a potential impact on the recommendation is not included in the set. As far as modelling assumptions used in the MCDA-MSS database are concerned, all methods that deal with a hierarchical structure of the set of criteria, accept uncertain criteria performances and/or incomplete criteria sets, are also assumed to handle a flat structure of the set of criteria, work with deterministic and/or complete criteria sets, respectively. Except for decision rules-based methods, when the user chooses an incomplete set of criteria, the MCDA-MSS points out that for the recommended methods, due to the aggregation of the performances on all the criteria in a single measure of quality, they should be used only when all relevant criteria are accounted for.

#### *2.1.1.2 MCDA-MSS section 2: Preference model (c.2)*

The definition of the preference model is the second section of the taxonomy in the MCDA-MSS. It consists of nine main features, starting with looking at the measurement scale used by the method (c.2.1). This can be of three types, qualitative, quantitative, and relative (Pap et al. 2009). The first one only considers the order of performances, the second accounts for the differences between performances, and the third is grounded on comparisons between alternatives to express preference intensity on a ratio scale. The methods that use scales in a quantitative matter are further divided according to how they assess the performance of the alternatives, into performance-based and pairwise comparison-based. In the former case, the methods evaluate the desirability of each alternative's performance individually (e.g., in MAVT (Keeney and Raiffa 1976)), while in the latter, the methods compare the alternatives pairwise to define if one performs at least as well as (or better than) another one (e.g., in ELECTRE methods (Figueira et al. 2016)). Performance-based methods that employ quantitative scales are moreover split according to how they treat the raw information before the aggregation step, distinguishing between those applying linear, piecewise linear, and non-linear transformation strategies (Nardo et al. 2008, Cinelli et al. 2020b). A subdivision is also applied to the pairwise comparison-based methods, discerning those that use comparisons of performance differences with (e.g., PROMETHEE methods (Brans and De Smet 2016)) and without thresholds (e.g., EVAMIX (Voogd 1982)). Lastly, the taxonomy distinguishes between the methods which use relative comparisons of performances to express the intensity of preference either in ordinal (e.g., MACBETH (Bana E Costa and Vansnick 1999)) or in ratio terms. The latter group further differentiates the methods according to whether they are performance-based (e.g., weighted means without transformation of criteria performances (Itsubo 2015)) or pairwise comparison-based (e.g., AHP (Saaty 1980)).

The second feature of the preference model is the one that considers how the comparison of the performances on the criteria is performed by the method. (c.2.2). Four options are provided with this feature:

1. Performances are transformed with a data-driven normalization approach and then compared (e.g., Nardo et al. (2008), Cinelli et al. (2020b));
2. Performances are compared by the DM with respect to the graded intensity of preference: The comparisons are performed by the DM who has to choose one value from a scale that is pre-defined (e.g., Saaty (1980), Bana E Costa and Vansnick (1999));
3. Performances are compared by the DM with respect to the non-graded intensity of preference: The comparisons are performed by the DM whose intensity of preference is not pre-defined in a set of values (e.g., Morton (2018), Grabisch and Labreuche (2005), Siskos et al. (2016));

4. Raw performances are compared directly (e.g., Greco et al. (2016b), Figueira et al. (2016).

The four features that follow define binarily whether standard components of MCDA methods are part of or not part of the DMP, being (i) the weights of the criteria (c.2.3), (ii) the per-criterion pairwise comparison thresholds (c.2.4), (iii) the interactions between criteria (c.2.5), and (iv) the multi-criteria profiles (c.2.6). Criteria weights are used to differentiate the importance of criteria in the aggregation procedure (Greco et al. 2019); per-criterion pairwise comparison thresholds characterize the preference sensitivity of the DM when comparing two alternatives on a single criterion (Granata 2017), interactions denote interdependencies between the criteria (Grabisch and Labreuche 2008), and multi-criteria profiles (Dias and Mousseau 2018) - not corresponding to the considered alternatives - serve as the basis for deriving the decision recommendation by comparing the performances of the alternatives with them.

Compensation level between criteria (c.2.7) is the seventh characteristic of the preference model, looking at how much the good performance on a criterion can compensate for the poor performance on another criterion (Rowley et al. 2012). Methods are assigned to one or more of the three available compensatory levels, which are null, partial, and full (Langhans et al. 2014).

The eighth feature is focused on determining whether and how the aggregation of the performances on multiple criteria should be performed by the method (c.2.8). In case of no aggregation, the respective method will develop the recommendation by considering the evaluation of the alternatives on a criterion-by-criterion basis (Bouyssou et al. 2006a). In case the aggregation takes place, three options are available, being scoring functions, binary relations, and decision rules (Słowiński et al. 2002). Scoring functions aggregate the individual criteria performances (usually normalized) to define the overall quality of each alternative (Nardo et al. 2008). Methods using binary relations employ pairwise comparisons of alternatives, which can be properly exploited to lead to a comprehensive assessment of each alternative (Roy 2016b). Lastly, rules-based methods aggregate the performances on different criteria using information connectors in the form of "if …, then …" statements (Greco et al. 2016b). These aggregation options are not mutually exclusive, and the analyst can request a method that employs one or more of these aggregation modes.

The last feature for the definition of the preference model is the capacity to deal with inconsistent preference information (c.2.9). Three types of inconsistencies can be considered, including (i) only violation of dominance relation (e.g., DRSA-based method (Greco et al. 2016b)), (ii) only other types of inconsistency, not including dominance (e.g., Best Worst Method (Rezaei 2015) and AHP (Saaty 1980)), and (iii) violation of dominance and other types of inconsistency (e.g., NAROR-Choquet (Angilella et al. 2010)). The case of violation of dominance only is further split between a strict and a relaxed treatment of the violation. The former captures the inconsistency of all objects (individual alternatives or pairs) violating the dominance principle (Greco et al. 2001a). In contrast, the latter captures the inconsistency of all objects (individual alternatives or pairs) insufficiently consistent (defined by a user-specified threshold on the consistency measure) with respect to the dominance principle (Greco et al. 2001b). The category including only other types of inconsistency and not including dominance accepts two sub-categories. The first one consists in handling inconsistency with respect to relative comparisons expressed on a cardinal scale for different pairs of alternatives and/or criteria (i.e., cardinal inconsistency) (Saaty 1980). The second refers to the capacity of handling inconsistency with respect to relative comparisons expressed on a cardinal scale for the same pair of criteria (i.e., inconsistency concerning preference comparison of criteria) (Rezaei 2015). The most comprehensive set of inconsistencies is the third one, as it captures the inconsistency of all pieces of preference information that violate the dominance principle and/or cannot be reproduced with an assumed (score- or relation-based) preference model.

#### *2.1.1.3 MCDA-MSS section 3: Elicitation of preferences (c.3)*

The type (c.3.1), frequency (c.3.2), and confidence (c.3.3) of preferences provision constitute the third section of the taxonomy. The first main distinction is between the type of preferences, which can either be direct or indirect (Dias et al. 2018). In case they are direct (c.3.1.1), the model parameters are defined directly, while in the indirect mode (c.3.1.2), local or holistic judgments of the experts/DMs on some reference alternatives are used to elicit them. With respect to direct preferences, the analyst can define whether the features are either specified directly by the decision maker or not. The latter includes

DMPs where the model parameters are missing, they cannot be specified, or the decision maker requires a method that does not use them.

Direct preferences (c.3.1.1) include four characteristics, starting from the weights of the criteria (c.3.1.1.1), which can be used to set the difference of importance between them (Greco et al. 2019). Two distinctive overarching categories can be defined for the type of criteria weights, namely precise and imprecise. As part of the precise ones, trade-offs and importance coefficients are distinguished (Munda and Nardo 2005). Trade-offs indicate the exchange rate accepted between the criteria to compensate each other performance (Munda 2008b). Relative importance coefficients specify the strength of one criterion in comparison with others in a voting-line procedure (Munda and Nardo 2005). Weights as precise relative importance coefficients are further split in those defined per-criterion (Riabacke et al. 2012) and with pairwise comparison ratios (e.g., the AHP method (Saaty 1980)).

In the case of imprecise weights, where no exact values are defined and they are driven by constraints of a different type, multiple options can be chosen, including missing input (Lahdelma et al. 1998), fuzzy numbers (Nădăban et al. 2016), ordering of some or all criteria with or without intensity of preferences (Dias and Climaco 2000, Punkka and Salo 2013), pairwise comparison-based difference of importance (Angilella et al. 2010), ratios (Salo and Hämäläinen 1992), distributions (Pelissari et al. 2020), intervals (Ahn 2017), and rank requirements (Salo and Punkka 2005). The distinction between absolute comparisons (i.e., per-criterion) and pairwise comparison ratios is also applied. With respect to the assumptions made in the development of the MCDA-MSS database, the methods that accept imprecise weights are also assumed to work with the weights in their precise form (e.g., SMAA-PROMETHEE II (Corrente et al. 2014) can work with imprecise weights defined with an ordering of some of the criteria, but it can also accept precise relative importance coefficients).

Pairwise comparison thresholds (c.3.1.1.2) are the second characteristic of direct preferences. They can be used to characterize the preference sensitivity of the DM when comparing two alternatives. The three most common types are included, being indifference, preference, and veto thresholds (Figueira et al. 2016), each distinguished according to whether they are precise or imprecise. Indifference threshold is the maximum difference between performances of two alternatives under which they are considered indifferent (e.g., if the price of two cars differ by not more than $100, then they are indifferent on the price) (Dias and Mousseau 2018). Preference threshold sets the minimum difference between performances of two alternatives above which a strict preference can be defined (e.g., if the price of two cars differs by at least $2,000, then one strictly prefers the cheaper option) (Dias and Mousseau 2018). Veto threshold determines the minimum difference in performances of two alternatives which, when exceeded, invalids the preference of the worse alternative over the better one, irrespective of their performances on the remaining criteria (e.g., if one car costs more than $10,000 in comparison to another car, then the former cannot be preferred to it at the comprehensive level despite its advantages on the remaining criteria) (Dias and Mousseau 2006). It was assumed that methods using these imprecise pairwise comparison thresholds could also accept precise ones. Even though we explicitly consider only three types of thresholds, more do exist more (e.g., discordance, reinforced preference, or counter-veto) (Roy and Słowiński 2008). When the use of these others in MCDA becomes more common, the taxonomy can be extended to account for each of them explicitly.

The third feature of direct preferences is the type of interactions between the criteria (c.3.1.1.3), which are divided into two groups, (i) positive and negative, and (ii) antagonistic. Positive interaction means that the comprehensive impact of the criteria on the quality of an alternative is greater than the impact of these criteria taken separately (Grabisch and Labreuche 2008). Negative interaction implies that the comprehensive impact of the criteria on the quality of an alternative is smaller than the impact of these criteria taken separately (Grabisch and Labreuche 2008). The antagonistic effect operates by lowering the influence of other criteria in the case the performance on one criterion is very low (Figueira et al. 2009a). All these interactions can be provided as precise and imprecise, according to how much knowledge is available on the DMP.

The last feature of the direct preferences is the definition of the discriminatory profiles (c.3.1.1.4) when sorting, choice, and ranking problems are chosen. This can be attained either by using characteristic profiles (i.e., representative/most typical for each class) (Almeida-Dias et al. 2010) or boundary profiles (i.e., DM-specified frontiers/boundaries which steer their assessment) (Figueira et al. 2016). The methods may incorporate single or multiple profiles to characterize each class or form the basis for comparing the existing alternatives (Fernández et al. 2017). What is more, profiles for sorting are differentiated between precise and imprecise (Fernández et al. 2019), using the modelling assumption that methods accepting the imprecise type also accept the precise one.

There exist other types of parameters whose values need to be specified to run a specific method and derive a recommendation. Such example parameters include a number of characteristic points of marginal value functions deciding upon their flexibility and ability for representing various decision-making policies (Jacquet-Lagreze and Siskos 1982), a credibility threshold indicating the minimal value of a valued outranking relation justifying the truth of a crisp relation (Figueira et al. 2016), or a compensation parameter between different types of aggregation functions that are jointly used to evaluate decision alternatives (Brauers and Zavadskas 2010). However, they cannot be categorized under any more general type of characteristic.

Direct specification of model parameters is sometimes a challenging task for the DM, who might have difficulties in understanding their meaning and/or not have the time to devote to providing input on each of them. For these reasons, indirect elicitation techniques (c.3.1.2) that use local or holistic judgments of the experts/DMs on some reference alternatives have become popular in the last decades (Jacquet-Lagrèze and Siskos 2001, Doumpos and Zopounidis 2011). The reasons are that experts/DMs find themselves more comfortable in exerting choices rather than explaining them (Buchanan and O'Connell 2006).

As far as sorting problems are concerned, six types of indirect preferences are offered. The first is the assignment of reference alternatives to decision classes (Özpeynirci et al. 2018), while the second one accounts for such assignment but with a variable level of certainty (Liu et al. 2020b). The third type is the assignment-based pairwise comparisons of alternatives (e.g., alternative A should be assigned to a decision class at least as good as alternative B) (Özpeynirci et al. 2018). And the fourth is the desired comprehensive value of alternatives assigned to a given class or class range (e.g., alternatives assigned to a class at most medium should have a value not greater than 0.4) (Kadziński et al. 2015). The last two options include the comparisons of reference alternatives with respect to intensity of preference expressed on an ordinal (Ishizaka and Gordon 2017) or ratio scale (Ishizaka et al. 2012), respectively.

Regarding ranking and choice problems, five types of indirect preferences are considered. The pairwise comparisons of the alternatives (e.g., alternative A is preferred to alternative B) represents the first type. The second consists in the comparisons of reference alternatives with respect to the intensity of preference expressed on an ordinal scale (e.g., alternative A is very strongly preferred to alternative B) (Argyris et al. 2014), followed by the comparisons of reference alternatives with respect to the intensity of preference expressed on a ratio scale (e.g., alternative A is preferred to alternative B by a factor of 2 on criterion 1) (Saaty 1980). The set is completed by the rank-related requirement (e.g., alternative A is in the top 3) (Kadziński et al. 2013) and by the desired comprehensive value of alternatives (e.g., the value of the alternative with at least the fifth rank is at least 0.6) (Salo and Hamalainen 2001).

The second to last feature of the preference model is the frequency of preference input (c.3.2). It distinguishes whether the preferences are provided at the start of the elicitation process or whether they are given successively in different iterations (Tsoukiàs 2007), thus encouraging the user to observe the evolution or convergence of the recommendation.

The elicitation of preferences is completed by accounting for the level of confidence in the provided preferences (c.3.3), offering the option of distinguishing between preferences with or without a confidence level, with such levels that can be, e.g., characterized by the statements ‘‘absolutely sure”, ‘‘sure”, or ‘‘mildly sure” (Greco et al. 2010). The modelling assumption used here for the MCDA-MSS database development is that methods working with a level of confidence can also work without it.

#### *2.1.1.4 MCDA-MSS section 4: Exploitation of the preference model (c.4)*

The last section of the taxonomy aims to delineate how the preference model should be exploited to derive or enhance the decision recommendation. Two main options are available, with a univocal recommendation (c.4.1) on one side, and the output variability analysis (c.4.2) on the other side. The first consists in offering a univocal outcome for the chosen problem statement. The second focuses on showing how variable the recommendation can be when there is uncertainty with respect to the performances of alternatives and/or representation of the DM's preferences by an assumed preference model (Dias 2007).

Univocal recommendations can be derived with or without output variability analysis. When no output variability analysis is used (c.4.1.1), the recommendation can be provided either by a single or multiple contingent models (Kadziński et al. 2020). The option of a single model is further differentiated in deterministic and representative. The deterministic model is a single preference model instance with precise parameter values directly specified by the DM (Nardo et al. 2008). The representative model is a single preference model instance with precise parameter values that are either selected by the method (i.e., algorithmic) (Jacquet-Lagreze and Siskos 1982) or directly chosen by the DM (i.e., direct involvement) (in both cases, there exist other possible preference model instances) (Kostkowski and Słowiński 1996).

When output variability analysis is embedded in the development of the univocal recommendation (c.4.1.2), the taxonomy (and the MCDA-MSS) offers two options. The first is compromise exploitation, where the univocal recommendation is constructed by aggregating or building on the outcomes of output variability analysis without selecting a representative preference model instance (Vetschera 2017). The second is representative exploitation, which derives the univocal recommendation from an application of a representative preference model instance selected based on the outcomes of output variability analysis (Greco et al. 2011, Kadziński et al. 2012b).

The last part of the taxonomy is devoted to defining how to conduct the output variability analysis (c.4.2.1) and which results to focus on (c.4.2.2). Two types of analysis (c.4.2.1) can be performed, with the first providing the extreme results with all compatible models (Corrente et al. 2013) and the second supplying the distribution of results with a sample of compatible models (Lahdelma and Salminen 2010). Regarding what results to focus on (c.4.2.2), a distinction is made between choice, ranking, and sorting problems. For choice and ranking problems, the exploitation of the model can be tailored to the selection of the alternatives (with, e.g., robustness analysis on the kernel and/or preference relations (Govindan et al. 2019)), the ranking (with, e.g., extreme ranking analysis (Kadziński et al. 2012a)), the score (with, e.g., score variability (Dias and Climaco 2000)), the preference relations and also preference intensities (Figueira et al. 2009b). For sorting problems, the exploitation of the model is available for class assignments (e.g., alternative A is assigned to class 2 with all the models (Greco et al. 2010)), assignment based pairwise relations (e.g., alternative A is assigned to a class at least as good as alternative B for all models) (Kadziński et al. 2015), and class cardinalities (e.g., at least ten and at most 15 alternatives can be assigned to class medium (Kadziński et al. 2015)).

***Table 1***: *Taxonomy of the MCDA-MSS features. See bottom of the table for acronyms.*

### *2.1.2 The database of MCDA methods used in the MCDA-MSS*

The database of 205 MCDA methods in the MCDA-MSS is, to the best of the authors' knowledge, the widest available of the DSSs that have been proposed to recommend such methods. It comprises representatives of the three main families of MCDA methods, namely approaches incorporating scoring functions, binary relations, and decision rules. As far as the methods employing scoring functions are concerned, there are widely used representatives like AHP (Saaty 1980), ANP (Saaty 2016), MAVT (Keeney and Raiffa 1976), TOPSIS (Hwang and Yoon 1981), VIKOR (Opricovic and Tzeng 2004), and additive weighted average (Razmak and Aouni 2014, Cajot et al. 2017, Thies et al. 2019). Less common methods like EVAMIX (Voogd 1982), and REMBRANDT (Van den Honert and Lootsma 2000) complement this set. A similar trend is visible for the binary relation methods, where several original options of the most used from this family were included, namely those from ELECTRE and PROMETHEE ones (Behzadian et al. 2010, Govindan and Jepsen 2016). In this case, too, less frequently used methods are incorporated, such as ARGUS (De Keyser and Peeters 1994), NAIADE (Munda 1995), QUALIFLEX (Paelinck 1976), and REGIME (Hinloopen et al. 1983). Regarding methods based on decision rules, this led to the inclusion of Dominance-based Rough Sets Approach (DRSA) methods (Greco et al. 2016b), which have never been accounted for in this type of DSSs.

A broad set of methods that include dated as well as recent developments in the MCDA area was then selected to complement the dataset. The rationale for their selection was as follows:

1. Expansion of ELECTRE and PROMETHEE methods to account for the advancement of these well-known methods to deal with more elaborate DMPs, like ELECTRE TRI-nB (Fernández et al. 2017), ELECTRE TRI-nC (Almeida-Dias et al. 2012), ELECTRE SORT (Ishizaka and Nemery 2014), PROMETHEE$^{GKS}$ (Kadziński et al. 2012a), and PROMSORT (Araz and Ozkarahan 2007);

2. Methods dealing with less common problem types including multiple criteria clustering (e.g., P2CLUST (Smet 2013), multiple criteria sorting with unknown decision classes (Rocha et al. 2013)), sorting with partially ordered decision classes (e.g., ELECTRE-SORT (Ishizaka and Nemery 2014)), and problems with decision classes size constraints (e.g., sorting with constrained decision classes (Özpeynirci et al. 2018), constrained choice (Podinovski 2010);
3. Methods that deal with a hierarchical structure of a family of criteria such as ELECTRE III-H (Del Vasto-Terrientes et al. 2015) and MCHP-UTA (Corrente et al. 2012);
4. MCDA methods that work with imprecise and indirect preferences, in order to include a very relevant and increasing branch in the MCDA domain (Doumpos and Zopounidis 2011, Corrente et al. 2013), which most DSSs have not accounted for, e.g., ACUTA (Bous et al. 2010), Preference programming with incomplete ordinal information (Punkka and Salo 2013), PRIME (Salo and Hamalainen 2001), DIS-CARD (Kadziński and Słowiński 2013), GRIP (Figueira et al. 2009b), UTA$^{GMS}$ (Greco et al. 2008), UTA$^{GMS}$ with imprecise evaluations (Corrente et al. 2017);
5. Single score methods tailored to problem formulations with, e.g., interactions (Grabisch and Labreuche 2016, Liu et al. 2020a), adjustable compensation levels (Cabello et al. 2018), hierarchical criteria, uncertainty, and criteria interactions (Angilella et al. 2018);
6. Methods that account for variability in the evaluation of performances on the criteria and/or preferences of the stakeholders while accounting for a sample of the compatible preference models, e.g., SMAA-ELECTRE-I (Govindan et al. 2019), SMAA-AHP (Durbach et al. 2014), or while exploiting the consequences of applying all compatible models with mathematical programming techniques, e.g., ARIADNE (Sage and White 1984), CUT (Argyris et al. 2014), or multiple criteria majority-rule sorting (Meyer and Olteanu 2019);

7. Less compensatory weighted sums that have been commonly used, especially in sustainability-related research (Langhans et al. 2014), i.e., geometric and harmonic.

Overall, 205 MCDA methods are included in the MCDA-MSS, mapped according to the taxonomy presented in Section 2.1.1. “Mapped” is interpreted in this research as the evaluation of which characteristics of the taxonomy are supported by the MCDA methods (see also Section 2.2 for more details). The complete database is available in Appendix B in the ESI.

## 2.2 Stage 2: Development of the MCDA-MSS in a web-software

The MCDA-MSS has been developed using rule-based modeling, which belongs to the first group of DSSs presented in Cinelli et al. (2020a). It is the most used modeling approach to develop this type of DSSs. Its added advantages and justifications for its selection for this DSS, are the objectivity, traceability, and understandability of the reasoning system empowered by the decision rules (Słowiński et al. 2009). This modeling uses rules in the form of *“if conjunction of conditions on some features is true, then decision is …”*. The database of MCDA methods already represents an information table shaped with this modeling approach, where the (sub-)characteristics are the conditions to describe the methods and the methods themselves are the decisions. Table 2 shows a sample of MCDA methods in the MCDA-MSS database, mapped according to a subset of features of the taxonomy for each type of problem statement. Once applied to the taxonomy of the MCDA-MSS, the rules syntax reads as “*If (sub-)characteristics a, b, … are activated/met by the DMP, then MCDA method I (, II, …) fits with the problem and can be recommended*”. Four examples of decision rules from Table 2 are as follows, with the 1s in the table showing those that are used to activate the rules:

1. “If the DMP is one requiring a partial and cardinal ranking, then GRIP method (Figueira et al. 2009b) is a suitable one”;
2. “If the DMP is one requiring a complete and cardinal order of decisions classes, with constraints on the number of alternatives assigned to the decision classes, then ROR-UTADIS (Kadziński et al. 2015) is a suitable method*”;*

3. “If the DMP is one requiring a complete clustering, then MCUC-CSA method (Rocha et al. 2013) is a suitable one”;
4. “If the DMP is one requiring a choice without any constraints on the number of alternatives to be recommended, then ELECTRE I method (Figueira et al. 2016) is a suitable one”.

***Table 2:*** *Example of MCDA methods in the MCDA-MSS database, mapped according to a subset of features of the taxonomy for each type of problem statement. Bold “1” are those that are used to trigger the rules in the main text.*

The DSS presented in this paper can be defined as a supporting tool for choosing the most appropriate MCDA method(s) for a given multiple criteria problem. It can substantially enhance, while selecting the relevant MCDA method(s), the interaction between the analyst and the DM in several ways.

First, the DMP is partitioned into four manageable sections. In each of these sections, a wide amount of information on the MCDA methods is neatly structured according to the taxonomy of (sub-)characteristics, using a sequenced stepwise questioning process (*DSS highlight 1 in* Figure 2). The explanations provided for each question within the information boxes constitute knowledge transfer and a learning exercise for the DM. These boxes are solutions to the black box effect that many MCDA methods have been criticized for. In fact, MCDA methods require several parameters to be defined to use them, and providing a brief description of each of them, can strengthen the confidence of the DM in their answers. An example for the information box showing the definition of “sorting” as a problem statement is presented in Figure 3.

***Figure 3:*** *Example of information box showing the definition of “sorting” as a problem statement*

Second, more than 200 methods are included in the dataset of the MCDA-MSS, which represents a notable library of well-established, as well as more recent methods (*DSS highlight 2 in* Figure 2). While an initial, although advanced step, towards creating a consistently organized repository of these methods, it can be of use and benefit for all operational researchers and beyond (see also Section 4.1).

Third, even if the dataset of MCDA methods in the MCDA-MSS is wide and a broad range of DMPs can be covered (*DSS highlight 3 in* Figure 2), there can still be cases where no method completely matches the description of a DMP (see Figure 4). In this case, the MCDA-MSS is equipped with a recommender function that allows the user defining the binding features that must be satisfied (in this example, the method must have the capability of discovering the preference direction of the criteria performances), as shown in Figure 5. This strategy caters a solution to the lack of a fully matching method to the DMP by offering one (or more) that is (or are) as close as possible to the DMP[i]. Figure 6 shows a sample of the methods that can still be recommended for the DMP in Figure 4, together with the feature(s) that are not supported by each method. In this case, UTA-NM and UTA-NM-GRA are shown as suitable methods for the DMP, though they are not tailored for hierarchical criteria with uncertain performances (see column "Missed" in Figure 6). It is important to note that the recommended method(s) in this scenario (i.e., no complete match with the DMP) should not be blindly applied. Instead, the analyst should discuss the feature(s) that are missed and define the course of action. This can imply that (i) the DMP is re-discussed to assess whether a different formulation of the problem can fully fit with a method, or (ii) a new method is developed that completely satisfies the requirements of the DMP under consideration.

***Figure 4:*** *Example of a DMP where no method completely matches the description of the DMP. The bright blue box highlights the information that is shown when this happens, and the hyperlinked "click here" leads to the window in Figure 5*

***Figure 5:*** *Choice box to select the features that must be satisfied by the method(s) that will be as close as possible to the DMP*

***Figure 6:*** *A sample of the recommended set of methods for the DMP presented in Figure 4, showing the features that are not satisfied by the UTA-NM and UTA-NM-GRA methods*

Fourth, an opposite situation to the one described above can emerge when the user cannot answer all the questions, and a large set of methods is recommended. This might be caused by a lack of information on the DMP. If there is the possibility of extending the interaction with the DM, the analyst must decide which questions to focus on to reduce the methods set. The button "Most selective questions" shows the questions that (increasingly) minimize the maximal number of recommended MCDA methods irrespective of the answer provided by the user. An example is presented in Figure 7, which shows that for a DMP where one would only select 'ranking' as problem statement and 'hierarchical' as criteria structure, the question on the comparison of performances is the first one that leads to as few methods as possible. In case the user cannot answer this question, the next most selective is the one on the scale used by the method. This capability of the MCDA-MSS of showing the reduction of methods according to each answer provides an immediate understanding of how the available methods are reduced according to the user's choices (*DSS highlight 4 in* Figure 2). The analyst can, in fact, look at this piece of information to tailor the efforts of interaction with the DM. For example, assuming two questions could be answered from the DM after some discussion, it is more efficient to select the question that leads to the lowest number of methods once answered. This objective is eventually achieved with the presented feature.

---

[i] In case resources and competencies are available, another solution to this challenge is that the analyst develops a tailored method for the specific DMP.

***Figure 7:*** *Activation of the "Most selective questions" button, showing that for this DMP the question on the comparison of performances is the most selective, followed by the one scale used by the method*

## 3. The MCDA-MSS in action

This section describes stage 3 of the methodology in Figure 2, presenting the test of the MCDA-MSS on a set of case studies selected from the literature. This was performed to (i) evaluate the applicability of the MCDA-MSS to multiple DMPs for different real-life applications, and (ii) discuss the MCDA method(s) used in the case studies and that/those recommended by the MCDA-MSS. The set-up of the test of MCDA-MSS with the literature case studies is firstly given in Section 3.1. Then, the results of its application are presented and discussed in Section 3.2.

### 3.1 Set-up for the MCDA-MSS test with literature case studies

The case studies selected for the test of the MCDA-MSS included a set of MCDA applications from the peer-reviewed literature. MCDA methods have been used in various application areas (Razmak and Aouni 2014, Cajot et al. 2017, Cegan et al. 2017, Frazão et al. 2018). A group was chosen as the database for the test of the MCDA-MSS. These included airline service quality (Liou 2011), land remediation (Sparrevik et al. 2012), biofuels energy chains (Dias et al. 2016), urban regeneration (Ferretti and Degioanni 2017), supplier selection (Syarah Raudhatul and Firman 2019), land use suitability analysis (Qiu et al. 2017), and energy production planning (Haurant et al. 2011, Maxim 2014).

It must be noted that a case study is defined as "a DMP that is solved with an MCDA method". Thus, there can be more than one case study in a single journal paper, as is the case for Dias et al. (2016), in this test of the MCDA-MSS, where two case studies are identified. It must also be stressed that the objective of this test was not to have a large set of studies from the same application area to apply the MCDA-MSS to derive some general conclusions on its applicability and performance (see Section 4.2 on future research). Rather, it was intended as an initial evaluation of the. MCDA-MSS in terms of its capability to transparently lead the analyst through the selection of MCDA methods. In addition, it provides preliminary considerations from the choices made by analysts in past research on the chosen MCDA methods.

The taxonomy features are implemented in the MCDA-MSS as a set of questions that evaluate whether those activated in each DMP are matched with that/those of a (set of) MCDA method(s). Table 3 presents a few example questions that can be used to describe the MCDA methods and the case studies. The taxonomy can, in fact, be used to describe the case studies in the same manner as the MCDA methods.

***Table 3:** Example questions used to describe the MCDA methods and a case study (Adapted from Cinelli et al. (2020a))*

Table 4 shows a simplified example of the description of the MCDA methods, the case studies, the chosen method(s) in the case studies, and that/those recommended by the MCDA-MSS. It can be noticed that as far as literature case study 1 is concerned, the method chosen by the authors of the study is the same as the one recommended by MCDA-MSS, while this is not true for literature case study 2.

***Table 4:** Example of a description of MCDA methods and case studies according to a sample of features of the taxonomy. ✓in "Description of the MCDA methods" = the method supports this feature; ✓in "Description of the case studies" = the case study requests this feature. The last three columns show the method(s) chosen by the authors of the case study, the*

*recommendation of the MCDA-MSS with the relevant method(s) for each case study with complete match, and the one(s) with the missed features, respectively (Adapted from Cinelli et al. (2020a) and Wątróbski et al. (2019))*

### 3.2 Results of the application of the MCDA-MSS to literature case studies

Nine case studies were selected from the peer-reviewed literature to study the applicability of the MCDA-MSS and its decision support capabilities, using its questioning procedure presented in Section 3.1. The results are summarized in Table 5, which shows (i) the MCDA method that was chosen by the authors of the case study, (ii) the decision-making features that have been missed by the MCDA method chosen in the case study with respect to the case study description by the authors of this paper, and (iii) the recommended MCDA method(s) by the MCDA-MSS with a complete match of the decision-making features and (iv) with missed features.

The taxonomy of the MCDA-MSS has been capable of describing completely all the case studies (for the mapping of each case study, see Appendix C in the ESI). The selected MCDA methods in six case studies are part of those recommended by the MCDA-MSS, as shown in Table 5. Based on the framing of the MCDA-MSS, this is an indication of the same description of the MCDA methods and the case studies, like in literature case study 1 in Table 4. Specific details on the choices driving the selection of the MCDA methods are provided in these cases. Ferretti and Degioanni (2017), who use a multi-attribute value function model, emphasize both the meaning of weights as tradeoffs and the need to create value functions to "harmonize" the different measurement scales. Dias et al. (2016) account for the robustness of the decision recommendation by adopting a value-based model with partial information on the input weights, showing the synergistic benefit of stochastic and exact robustness analysis. In the former (i.e., stochastic), the probability of receiving a certain rank provides a perspective on the trend of the results, while in the latter (i.e., exact), the visualization of how much one alternative can outperform the other highlights how different the alternatives do actually score. Liou (2011) devised a transparent and straightforward sorting system for airline service evaluation using the Dominance-based Rough Set Approach (Greco et al. 2016b), which has the additional noteworthy capacity of accepting inconsistencies in the input preferences. Haurant et al. (2011) stress the limited compensation, variable preference thresholds, and incommensurable measurement scales as the main reasons for the choice model they developed using ELECTRE IS (Figueira et al. 2016). Lastly, the added value of accounting for stochastic input was particularly emphasized in the work of Sparrevik et al. (2012), who used a ranking method (i.e., SMAA-PROMETHEE (Corrente et al. 2014)) to tackle a DMP that lacks deterministic performances, and provided a clear indication of robustness for the final recommendations.

***Table 5:*** *Summary of results of the MCDA-MSS test with the literature case studies*

However, the trend presented in the previous paragraph was not found in four case studies since the methods chosen by the authors of the studies were not included in those recommended by the MCDA-MSS. This implies that the description of the MCDA methods and the case studies are not the same, resulting in a situation like in literature case study 2 in Table 4. This set of case studies with a mismatch of MCDA methods shows how the MCDA-MSS can be a warning tool for errors committed in the selection of an MCDA method selection. The reason(s) for the mismatches are presented in the column "Missed features between the MCDA method chosen in the case study and the case study description by the authors of this paper" in Table 5.

Maxim (2014) proposed a composite indicator in the form of an index, obtained by combining min-max normalization and additive weighted mean, and it is the first mismatch found by the MCDA-MSS. The first reason for this mismatch is using an MCDA method (i.e., weighted mean) that is not suited for criteria whose measurement scale is ordinal, such as the ability to respond to demand and social acceptability. In these criteria, the numbers coding an order have qualitative meaning. Any operator that assumes that these numbers have a quantitative meaning implements compensations that are not mathematically justifiable, leading to results that are not scientifically meaningful (Ebert and Welsch 2004, Munda and Nardo 2009). The assumption of the quantitative meaning of these numbers can result in the following. The arbitrary choice of the coding in the input data can lead to different results, solely due to the discretionary numbering for the qualitative measurement scale (e.g., low = 1, medium = 2, high = 3 could also be represented as low = 2, medium = 4, high = 7 as far as the increase is preserved) (Pollesch and Dale 2016). The requirement of choosing an aggregation algorithm that can properly handle the input information is an important consideration that has been made repeatedly in the MCDA literature (in, e.g., Guitouni and Martel (1998), Roy (2016a)). The second reason for the mismatch is that the weights used in this case study were elicited with an elicitation protocol that provided importance coefficients as an outcome. Unfortunately, the weighted mean, which was the algorithm used by these authors, would have required weights in the form of trade-offs (Munda 2008a). The subjective assignment of importance weights based on expert judgment, irrespective of the measurement scale and with no reference to the acceptable trade-offs between the criteria, is a common mistake reported in the MCDA literature when a single score model is selected (Munda and Nardo 2005, Cinelli et al. 2014). The MCDA-MSS recommends QUALIFLEX (Paelinck 1976) as one of the methods as close as possible to this DMP, as summarized in the last column in Table 5. This is a method that uses weights as set in the study (i.e., importance coefficients). However, it implies that the stakeholder accepts that the way the information is used by the method consists in pairwise comparisons between the alternatives and not in taking into account the individual performance on each criterion.

The second mismatch was found in the case study of Syarah Raudhatul and Firman (2019). The MCDA-MSS shows this is also related to the process of weighting (see also Table 5), though in this case, it is driven by the use of the weights obtained with the AHP (Saaty 1980) in a method (PROMETHEE II (Brans et al. 1986)) that requires a different type of weights. AHP weights have a specific meaning assigned to them, the one of a relative priority of an element of the hierarchy which can be a (sub-)criterion or a performance of an alternative on a bottom-level (also called elementary) criterion. They serve to calculate the global score of each alternative as a sum of products of these weights along all paths of the hierarchy tree from the alternative to the goal. Each product includes the weight assigned to the performance of the alternative on a given criterion and the weights of all upper-level criteria on the path to the goal. Consequently, these are not trade-off weights, because they are not used in a weighted sum where the performances of alternatives on the bottom-level criteria are multiplied by the weights, and they are also not relative importance coefficients that are used in a voting-like procedure. These weights can be used meaningfully only in the AHP method. Their use in methods like PROMETHEE (Brans and De Smet 2016), ELECTRE (Figueira et al. 2016), or TOPSIS (Behzadian et al. 2012) is thus incorrect. Another interesting reason for the mismatch in this case study is the lack of inclusion of criteria interaction. In fact, one of the criteria is price, and its value is the same for each of the three chosen alternatives. This indicates that the inclusion of price as a criterion would be meaningful only if criteria interactions would have been included in the set. This aspect is considered by the MCDA-MSS, which recommends GAIA-SMAA-PROMETHEE-INT (GAIA-SMAA-PROMETHEE for hierarchical interacting criteria (Arcidiacono et al. 2018)) as the suitable method if the criteria weights are used as importance coefficients, and output variability analysis as an exploitation of the preference model is accepted by the user (which in the case of deterministic input, i.e., precise parameter values, will lead to a univocal recommendation).

The third mismatch was found by the MCDA-MSS is in the case study of Qiu et al. (2017), who used the weighted additive mean (Itsubo 2015), an MCDA method specifically developed for ranking DMPs and quantitative criteria, to tackle a sorting problem with qualitative criteria. The reasons for the erroneous use of weights from the AHP in another method and the use of number-coded qualitative criteria in a method interpreting all scales of criteria as quantitative have been presented above. The third aspect that causes the mismatch, as also summarized in Table 5, is a unique one in this test, and it relates to the formulation of the DMP. It shows the implication of a non-optimal formulation of the problem statement, a delicate and key step of the initial phase of any MCDA process (Bouyssou et al. 2006b). A weighted average is a method tailored to ranking problems, providing a score that is used to rank the alternatives. To obtain a sorting out of this score, arbitrary and abstract cut-off levels need to be defined to lead the assignments to different decision classes (e.g., 0.3 could be the threshold between a poor and medium overall performance). The MCDA-MSS suggests more suitable methods for this case study that involve a sorting problem statement, namely FlowSort (Nemery and Lamboray 2008) and PROMSORT (Araz and Ozkarahan 2005), assuming weights as importance coefficients.

Lastly, the MCDA-MSS signalled another mismatch for the case study by El Mazouri et al. (2018). The first justification is similar to the case study above since the authors applied a ranking method (ELECTRE III (Figueira et al. 2016)), to tackle a sorting problem. In fact, the authors aim at distinguishing the priority level to be assigned to different villages with respect to their need for electrification. This type of problem can be more appropriately managed with a sorting method rather than a ranking one, since the latter cannot explicitly define the level of priority assigned to each village. The second cause for the mismatch is the nature of the set of alternatives that are considered. The authors used ELECTRE III, which is tailored to problems with a stable set of alternatives. The problem under consideration is instead characterized by an increasing set of alternatives, since the villages to be evaluated by the model will keep growing on a supposedly regular basis, according to the budget availability. The final feature missed by the method selected by the authors is the type of set of criteria. Even if they state that more that more than 20 criteria should be considered, they only selected three, which makes the set incomplete. Based on the description of this DMP, MCDA-MSS recommends several methods that only miss the last feature (i.e., set of criteria = incomplete), indicating the need to study the decision recommendation with caution. A few examples of these methods are ELECTRE SORT (Ishizaka and Nemery 2014), OSMPOC (Nemery 2008), and THESEUS (Fernandez and Navarro 2011).

## 4. Conclusions

### 4.1 Key contributions of the MCDA-MSS

This paper presents a DSS, called the MCDA-MSS, that helps decision analysts describe complex decision-making processes and choose the MCDA method(s) relevant for each case study. It provides five main contributions.

First, it is a tool for decision analysts to transparently and comprehensively describe and reproducibly shape case studies while interacting with the DM. It can thus guarantee consistent and homogeneous communication among different stakeholders. The MCDA-MSS can streamline the complexity involved in MCDA methods selection, while being aware of the possibilities at each stage, as well as their implications. If this DSS would be used in future case studies, this could guarantee accountability and trust of the research, as well as enable its comparability. In fact, the MCDA-MSS provides a recordable description of the choices that led to selecting the MCDA method(s). It uses an accessible vocabulary to describe each step that needs to be taken while moving towards the method(s) selection. It also divides the selection process into four user-friendly sections, where a wide amount of information on the case studies is elicited and linked to the database of MCDA methods, using a stepwise questioning procedure *(DSS highlight 1)*.

Second, thanks to its extensive database of 205 MCDA methods and the large set of characteristics (156 in total), it is capable of dealing with DMPs that span from very simple to very complex, leading to the respective identification of those methods which are most appropriate *(DSS highlight 2)*. Suppose we adopt a knowledge transfer perspective among decision analysts. In that case, the taxonomy of the MCDA-MSS operationalized in its questioning procedure can also be seen as an educational and training tool for inexperienced decision analysts. They can first learn the components and steps necessary to lead an MCDA problem and then guarantee a sound description of each case study, resulting in an informed choice of the suitable MCDA method(s).

Third, it offers MCDA method(s) recommendations also for DMPs where there is no perfectly matching method. This feature ensures the analyst is never "left alone" in the selection process by making the most of the available information on the DMP and the knowledge in the MCDA-MSS *(DSS highlight 3)*.

Fourth, a refinement strategy for directing the user to the most discriminatory questions is proposed in cases where not all the questions can be answered. In these situations, the MCDA-MSS aids the decision analyst choose where to focus efforts to reduce the set of available options as efficiently as possible *(DSS highlight 4)*.

Fifth, the MCDA-MSS has the potentials to unveil methodological mistakes that analysts have made in a selection of the methods *(DSS highlight 5)*. These include the erroneous use of weights as importance coefficients instead of as trade-offs and the use of ordinal measurement scales in a method that requires quantitative ones. The use of the MCDA-MSS has the capacity to avoid these mistakes being committed in future case studies. Thus, making sure that the decision recommendation is aligned with the problem's characteristics and preferences of the involved stakeholders.

The MCDA-MSS was tested with a set of ten case studies from the peer-reviewed literature. It confirmed its capability to describe DMPs for different real-life applications, find suitable MCDA methods for each DMP, and unveil some methodological flaws in the selection of methods.

## 4.2 Future research

The MCDA-MSS and its underlying structure, called taxonomy, can represent an evolving repository of knowledge in the MCDA domain. In this respect, several extensions of this research can be envisioned. The first one consists in expanding the test of the MCDA-MSS by including a large set of case studies to identify trends in the use of MCDA methods. Another one could involve developing the (already substantial) pool of single DM-focused MCDA methods included in the MCDA-MSS, considering their ever-increasing availability. A further extension could be devoted to the inclusion of methods that support multiple DMs, an avenue of operational research that is crucial due to the recognition that many decisions are made not by individuals, but in groups, such as committees and boards (Lahdelma and Salminen 2001). Given the generality of the taxonomy used in the MCDA-MSS, it could also be possible to envision its adaptation to methods for Portfolio Decision Analysis (Liesiö and Vilkkumaa 2021). As far as software support is concerned, the presence of user-friendly software implementing MCDA methods can be of notable advantage in the development of the MCDA process and the interaction with the DM, making it a further very valuable feature for the extension of the MCDA-MSS.

The MCDA-MSS could also lead to recommending more than one method given its wide repository. This can lead to a notable challenge for the analysts, who might be asked to still propose one method for a certain project. This can be driven by the limited time availability or the DM’s commitment to providing one solution to the problem at hand. The final choice can then be influenced by some of the qualitative features of the methods, like the reported easiness of using it (Polatidis et al. 2006), its use in certain application areas (Moghaddam et al. 2011), or the availability of software implementing it (Weistroffer and Li 2016). One solution provided in this regard is to develop suitability indices to aggregate the (sub-)characteristics of the DMP and provide a synthetic measure of the suitability of different methods, which are respectively ranked according to it (e.g., Li et al. (2008) and Guarini et al.

(2018)). This, however, includes subjectivity due to the inherent nature of the qualitative features as well as their aggregation, making the final result less based on objective facts.

Overall, the MCDA-MSS is intended to help an analyst facing a multiple criteria decision problem to choose consciously an MCDA method that will respond positively to the needs of the DM and satisfy all the constraints characterizing the decision situation. The authors of this paper are eager to engage in a community-wide initiative involving experts in MCDA methods, decision analysts using these methods, and decision makers receiving decision recommendations. This combined action can result in an expansion of the methods repository as well as the tests on case studies, coupled with the inclusion of additional decision aiding features in the web-software. The outcome of this initiative can be a sustained contribution to the relevant and transparent use of MCDA methods to solve real-world problems. All these activities can be monitored on the MCDA-MSS dedicated webpage http://mcdamss.com.

# 5. Disclaimer

The U.S. Environmental Protection Agency, through its Office of Research and Development, collaborated in the research described herein as part of the Collaborative Research and Development Agreement #1040-18 with Poznan University of Technology. It has been subjected to the Agency's administrative review and has been approved for external publication. Any opinions expressed in this paper are those of the authors and do not necessarily reflect the views of the Agency. Therefore, no official endorsement should be inferred. Any mention of trade names or commercial products does not constitute endorsement or recommendation for use.

Each method in the database of the MCDA-MSS required the assignment of binary values (i.e., either 0 or 1) to each possible answer considered in the taxonomy. Given that the selected possible answers are 156, this means that the database of the MCDA-MSS is composed of 205 rows and 156 columns, requiring the manual assignment of 31,980 binary values (i.e., 1 or 0). Given the very large number of manual inputs, the authors acknowledge that there might be some mappings of MCDA methods that other authors/researchers might not agree with. Rather than this being cause for arguments between the developers of the MCDA-MSS and those authors, we encourage them to contact us to improve the database and contribute to this long-lasting initiative, since this is just the beginning of the MCDA-MSS.

Lastly, the developers of the MCDA-MSS do not take responsibility for the choices a user makes based on the recommendations of the MCDA-MSS.

# 6. Funding

Marco Cinelli acknowledges financial support from the European Union's Horizon 2020 research and innovation programme under the Marie Skłodowska-Curie grant agreement No 743553. Miłosz

Kadziński acknowledges financial support from the Polish National Science Center under the SONATA BIS project (grant no. DEC-2019/34/E/HS4/00045). Grzegorz Miebs acknowledges support from the Polish Ministry of Science and Higher Education under the Diamond Grant project (grant no. DI2018 004348). The research of Roman Słowiński was supported by the Polish Ministry of Education and Science (grant no. 0311/SBAD/0709).

# 7. Declaration of interests

The authors declare that they have no known competing financial interests or personal relationships that could have appeared to influence the work reported in this paper.

# Figures of the paper

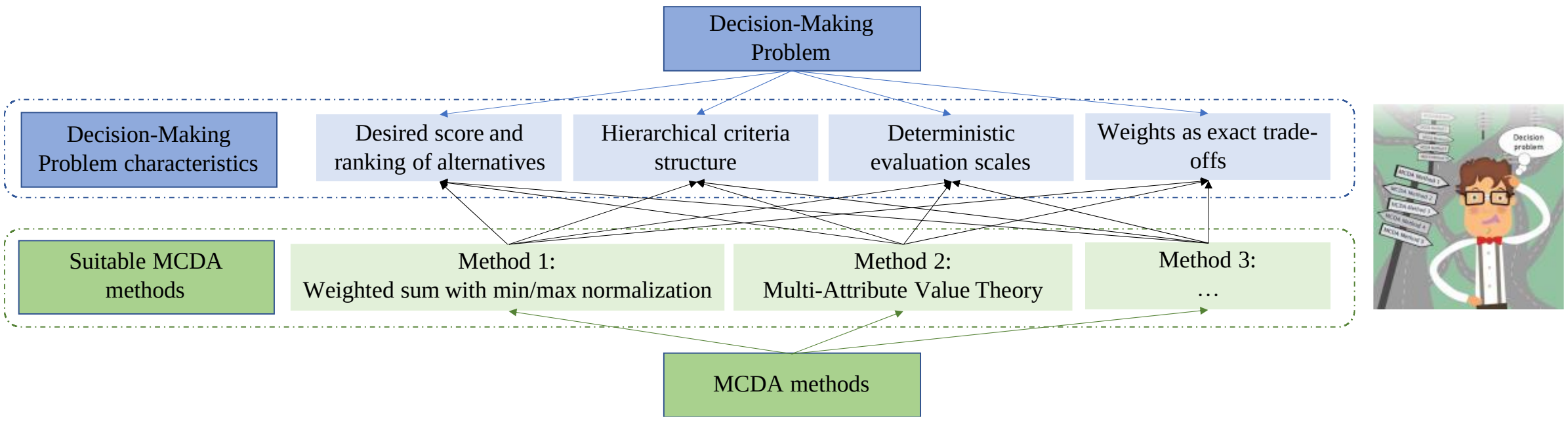

***Figure 1:*** *The challenge faced by decision analysts when the DMP is structured*

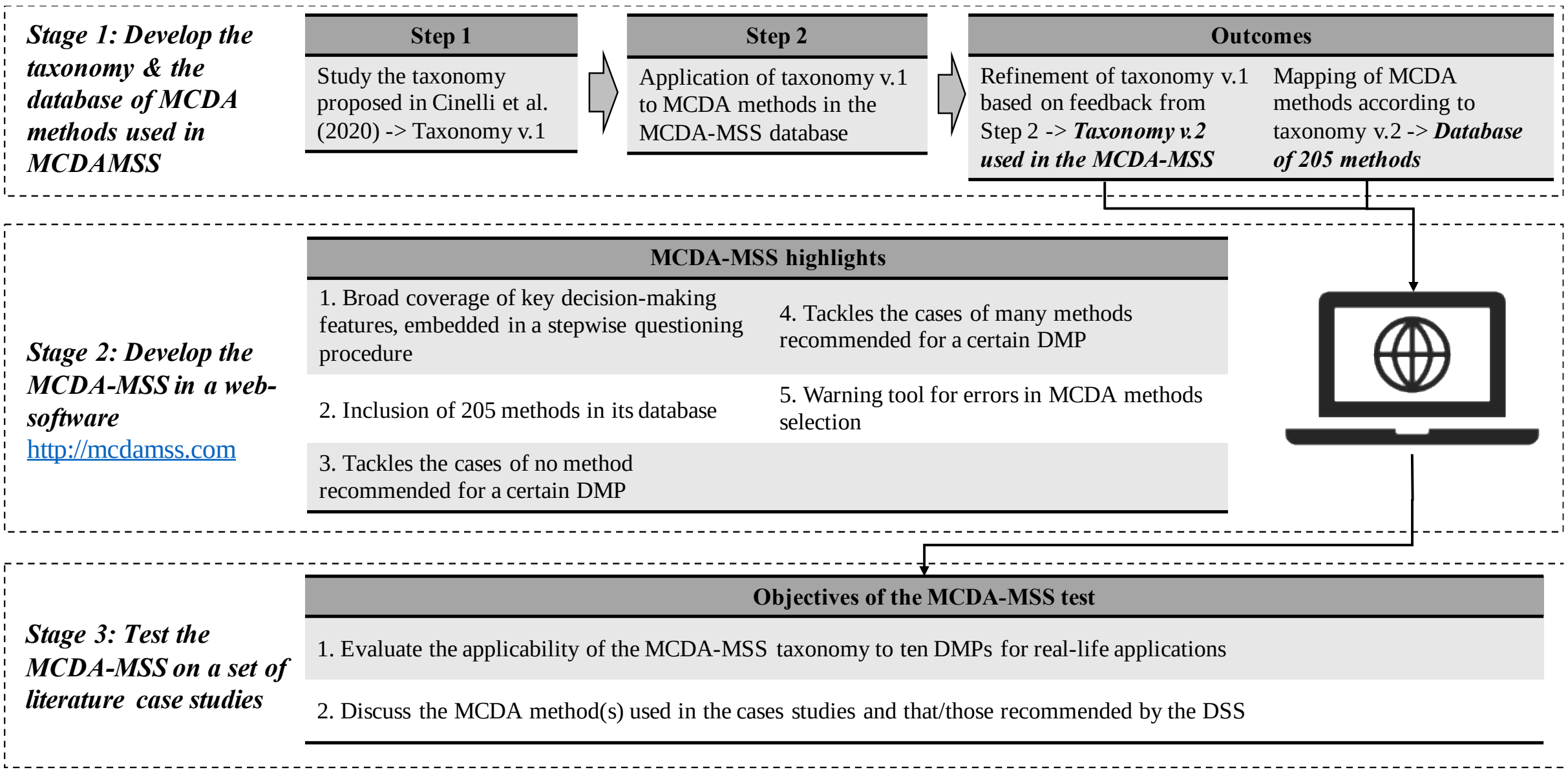

***Figure 2:*** *The methodology used to develop and test the MCDA-MSS*

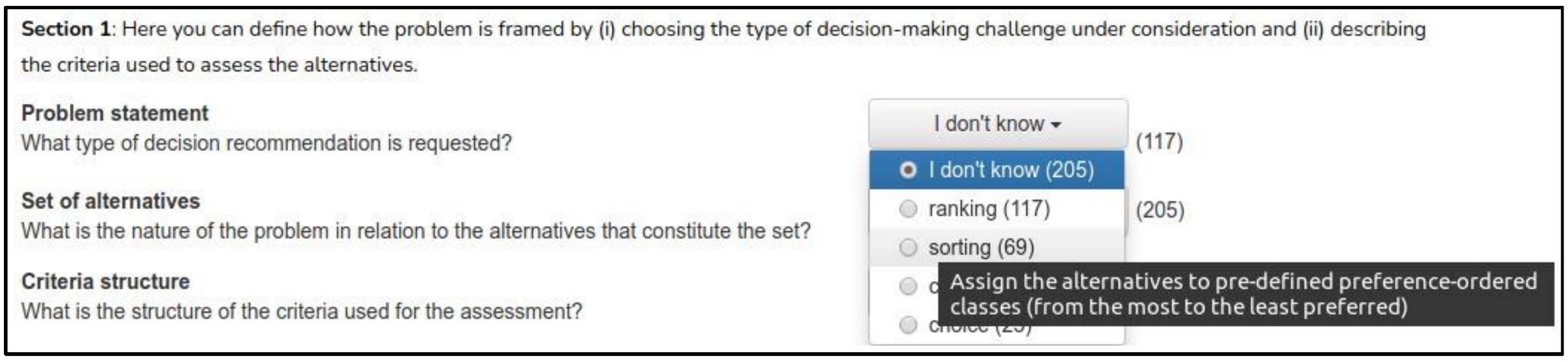

***Figure 3:*** *Example of information box showing the definition of "sorting" as a problem statement*

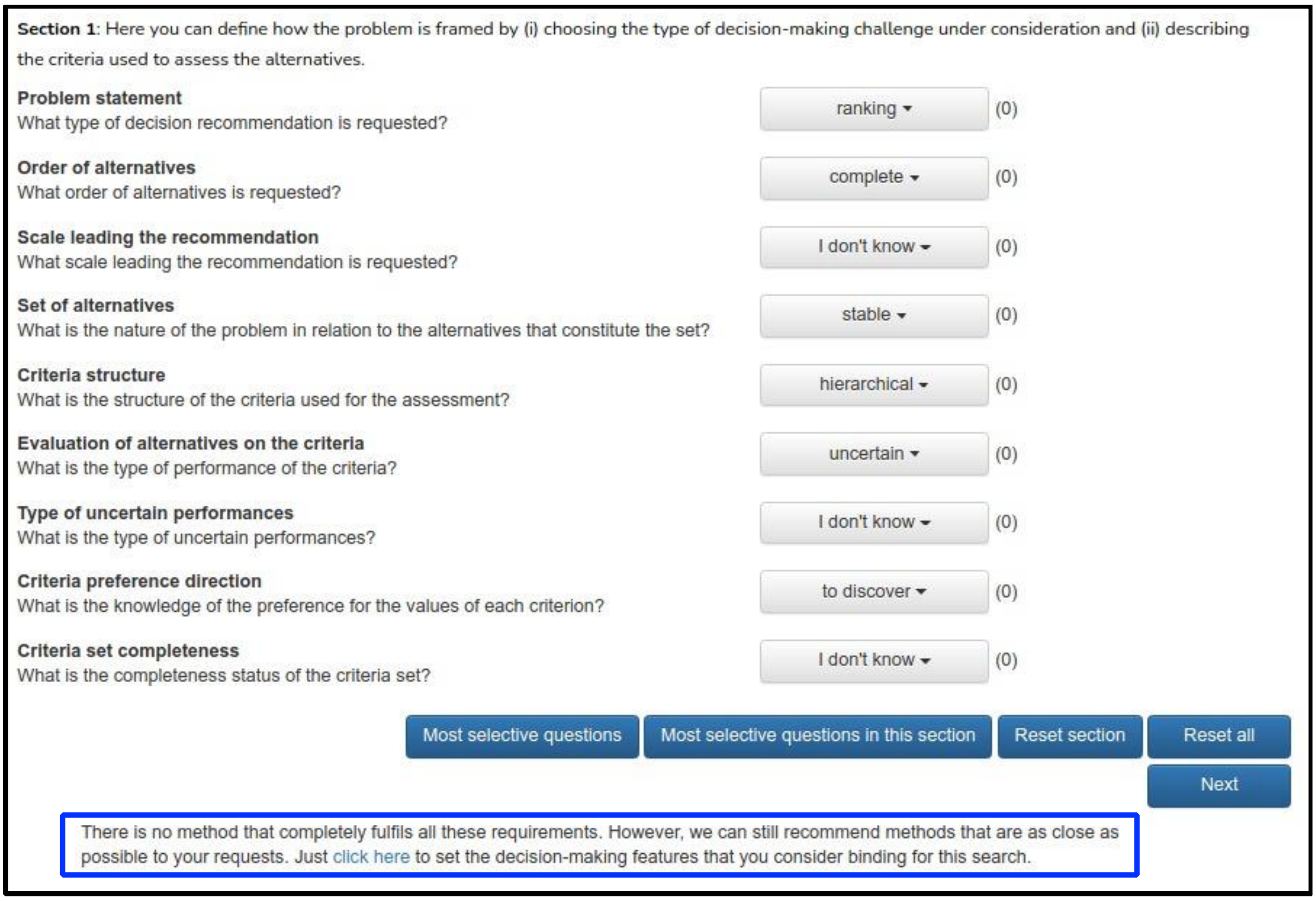

***Figure 4:*** *Example of a DMP where no method completely matches the description of the DMP. The bright blue box highlights the information that is shown when this happens, and the hyperlinked "click here" leads to the window in Figure 5*

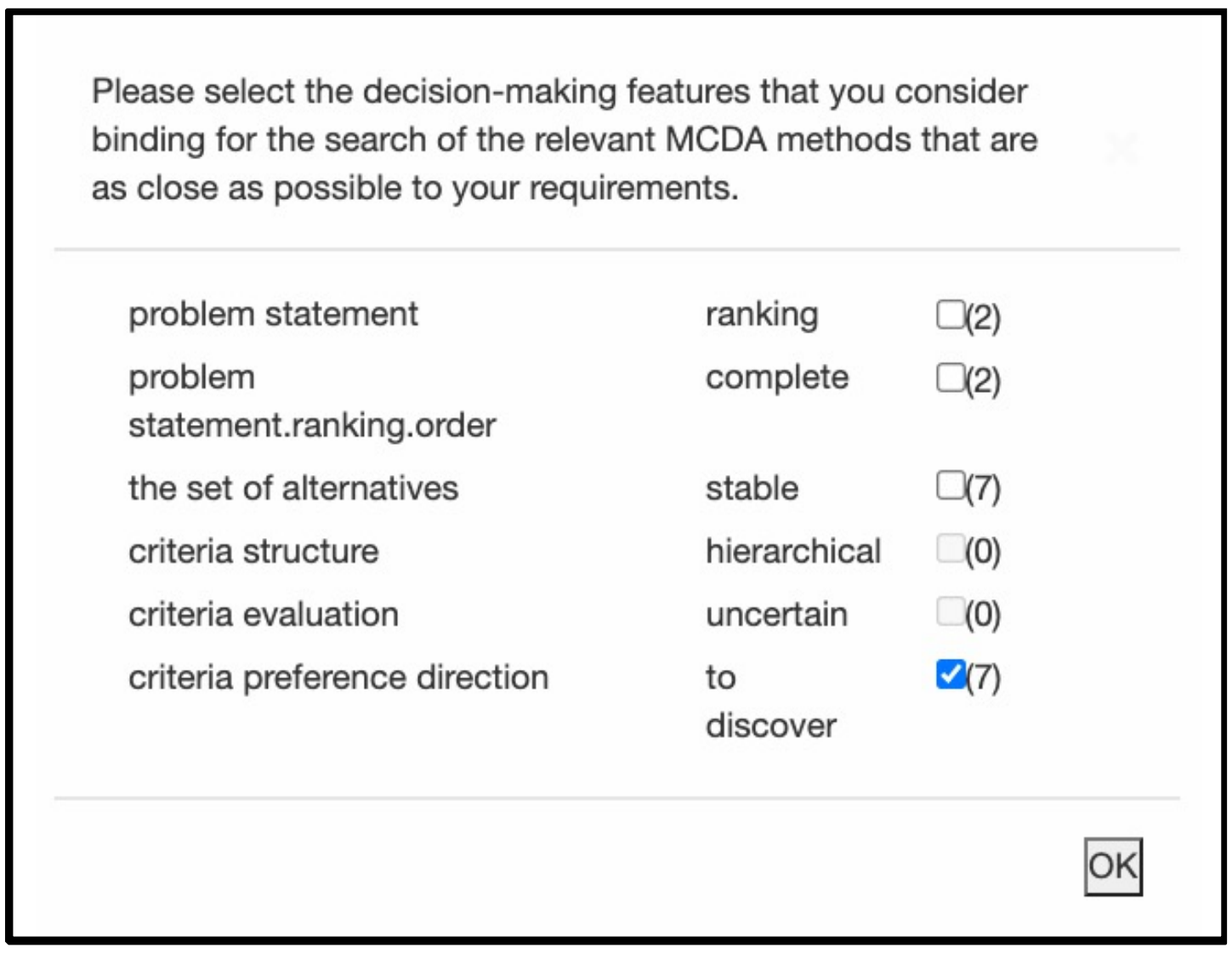

***Figure 5:*** *Choice box to select the features that must be satisfied by the method(s) that will be as close as possible to the DMP*

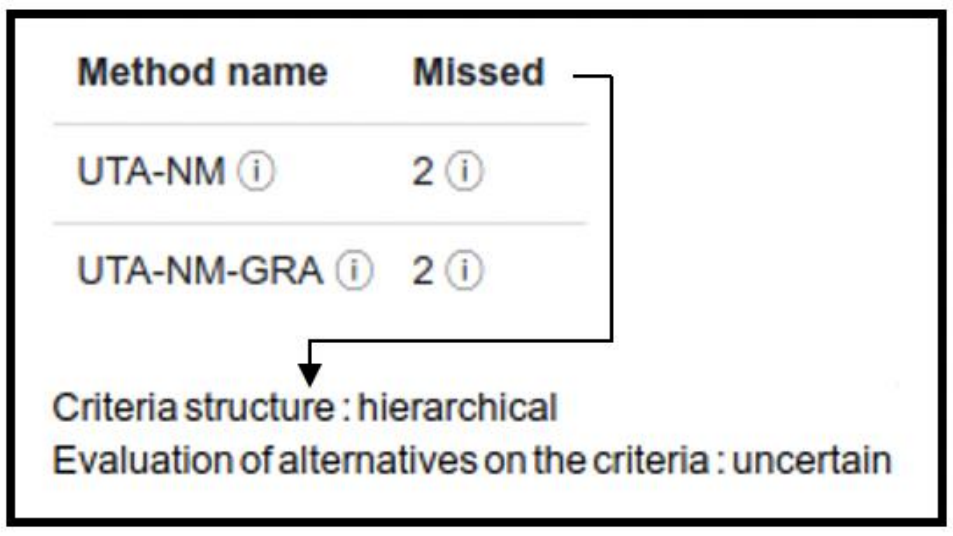

***Figure 6:*** *A sample of the recommended set of methods for the DMP presented in Figure 4, showing the features that are not satisfied by the UTA-NM and UTA-NM-GRA methods*

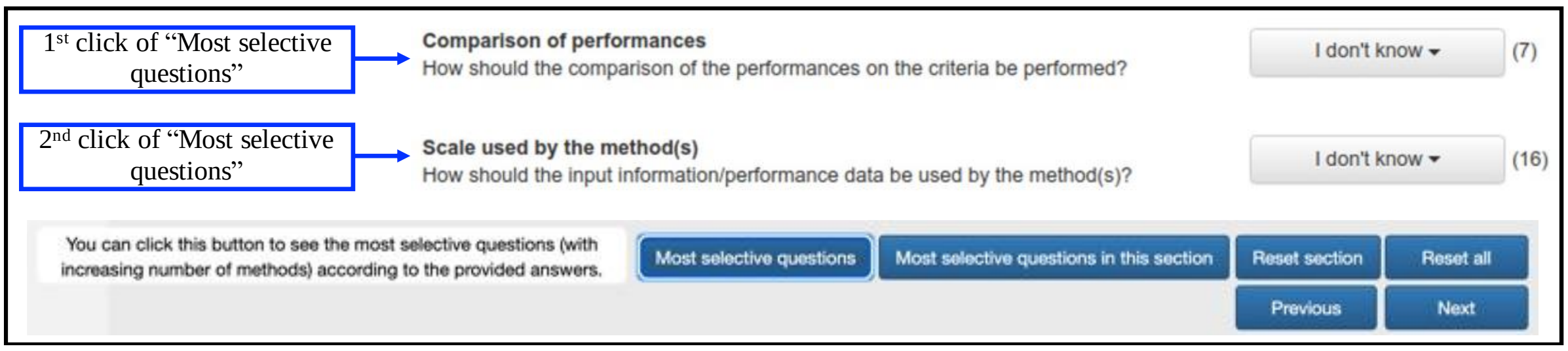

***Figure 7:*** *Activation of the "Most selective questions" button, showing that for this DMP the question on the comparison of performances is the most selective, followed by the one scale used by the method*

# Tables of the paper

***Table 1:*** *Taxonomy of the MCDA-MSS features. See bottom of the table for acronyms.*

| | | | | |
|---|---|---|---|---|
| **c.1 - Problem typology** | c.1.1 - Problem statement | Ranking | Order | Partial |
| | | | | Complete |
| | | | Scale leading the recommendation | Ordinal |
| | | | | Cardinal |
| | | Sorting | Order of decision classes | Partial |
| | | | | Complete |
| | | | Scale leading the recommendation | Ordinal |
| | | | | Cardinal |
| | | | Cardinality | With constraints |
| | | | | Without constraints |
| | | Clustering | No order | Non-relational |
| | | | | Relational: Partial tournament |
| | | | | Relational: Complete tournament |
| | | | Order | Partial |
| | | | | Complete |
| | | Choice | Cardinality | With constraints |
| | | | | Without constraints |
| | c.1.2 - The set of alternatives | Stable | | |
| | | Incremental | | |
| | | Flat | | |
| | c.1.3 - Criteria structure | Hierarchical | | |
| | c.1.4 - Evaluation of alternatives on the criteria | Deterministic | Per-alternative | |
| | | | Relative performance comparison | |
| | | Uncertain | Per-alternative | Missing |
| | | | | Interval |
| | | | | Probabilistic values |
| | | | | Fuzzy |
| | | | | Evidential reasoning |
| | | | Relative performance comparison | Interval |
| | | | | Fuzzy |
| | | | | Ratio imprecise |
| | c.1.5 - Criteria preference direction | Known | Monotonic | |
| | | | Non-monotonic | |
| | | To discover | | |
| | c.1.6 - Completeness of the criteria set | Complete | | |
| | | Incomplete | | |
| **c.2 - Preference model** | c.2.1 - How the input information/performance data is used by the method | Qualitatively | Performance-based | General function |
| | | | Pairwise comparison-based | Comparison of order of performances without thresholds |
| | | Quantitatively | Performance-based | Linear |
| | | | | Piecewise linear |
| | | | | Non linear |
| | | | Pairwise comparison-based | Comparison of differences with thresholds |
| | | | | Comparison of differences without thresholds |
| | | Relatively | Ordinal terms | Pairwise comparison-based |
| | | | Ratio terms | Performance-based |
| | | | | Pairwise comparison-based |
| | c.2.2 - Comparison of performances | Performances are transformed with a data driven normalization approach and then compared | | |
| | | Performances are compared by the DM with respect to graded intensity of preference | | |
| | | Performances are compared by the DM with respect to non graded intensity of preference | | |
| | | Raw performances are compared directly | | |
| | c.2.3 - Weights of the criteria | Yes | | |
| | | No | | |
| | c.2.4 - Per-criterion pairwise comparison thresholds | Yes | | |
| | | No | | |
| | c.2.5 - Interactions between criteria | Yes | | |
| | | No | | |
| | c.2.6 - Criteria profiles | Yes | | |
| | | No | | |
| | c.2.7 - Compensation level between criteria | None | | |
| | | Partial | | |

| | | | | | |
|---|---|---|---|---|---|
| | | Full | | | |
| | c.2.8 - Aggregation of multiple criteria evaluations | No | | | |
| | | Yes | Scoring function | | |
| | | | Binary relations | | |
| | | | Rules | | |
| | c.2.9 - MCDA method capacity to handle inconsistent preference information | Yes | Only violation of dominance | Only violation of dominance | Strict treatment |
| | | | | | Relaxed treatment |
| | | | Only other types of inconsistency, not including dominance | Only other types of inconsistency, not including dominance | Cardinal inconsistency |
| | | | | | Inconsistency with respect to preference comparison of criteria |
| | | | | | Other |
| | | | Violation of dominance and other types of inconsistency | Violation of dominance and other types of inconsistency | Value- and outranking-based preference disaggregation methods |
| | | | | | Other |
| | | No | | | |

| | | | | | | | |
|---|---|---|---|---|---|---|---|
| **c.3 - Elicitation of preferences** | c.3.1 - Type | c.3.1.1 - Direct | c.3.1.1.1 - Specify weights | Yes | Precise | Trade-offs | Per-criterion |
| | | | | | | Relative importance coefficients | Per-criterion |
| | | | | | | | Pairwise comparison ratio |
| | | | | | Imprecise - Per-criterion | FU | |
| | | | | | | ORAC | |
| | | | | | | ORCWP | |
| | | | | | | ORSC | |
| | | | | | | PWCD | |
| | | | | | | Ratio | |
| | | | | | | Distribution | |
| | | | | | | INT | |
| | | | | | | RRR | |
| | | | | | | Other constraints | |
| | | | | | Imprecise - Pairwise comparison ratio | FU | |
| | | | | | | INT | |
| | | | | No | | | |
| **c.3 - Elicitation of preferences** | c.3.1 - Type | c.3.1.1 - Direct | c.3.1.1.2 - Specify per-criterion pairwise comparison thresholds | Yes | Indifference | Precise | |
| | | | | | | Imprecise | |
| | | | | | Preference | Precise | |
| | | | | | | Imprecise | |
| | | | | | Veto | Precise | |
| | | | | | | Imprecise | |
| | | | | No | | | |
| | | | c.3.1.1.3 - Specify interactions between criteria | Yes | Positive, negative | Positive, negative - Precise | |
| | | | | | | Positive, negative - Imprecise: sign | |
| | | | | | | Positive, negative - Imprecise: intensity | |
| | | | | | Antagonistic | Antagonistic - Precise | |
| | | | | | | Antagonistic - Imprecise: only existence | |
| | | | | No | | | |
| | | | c.3.1.1.4 - Specify criteria profiles | Yes | For sorting | Characteristic profiles | Single: precise |
| | | | | | | | Single: imprecise |
| | | | | | | | Multiple: precise |
| | | | | | | | Multiple: imprecise |
| | | | | | | Boundary profiles | Single: precise |
| | | | | | | | Single: imprecise |
| | | | | | | | Multiple: precise |
| | | | | | | | Multiple: imprecise |

| | | | | | | |
|---|---|---|---|---|---|---|
| | | | | | For ranking or choice | General |
| | | | | No | | |
| **c.3 - Elicitation of preferences** | c.3.1 - Type | c.3.1.2 - Indirect | Sorting problems | AAC | At the comprehensive level | Precise |
| | | | | | | Imprecise |
| | | | | | At the lower level in the hierarchy | Precise |
| | | | | | | Imprecise |
| | | | | VAC | | |
| | | | | APC | | |
| | | | | DCVC | | |
| | | | | CRAIO | | |
| | | | | CRAIR | At the global level | |
| | | | | | At the local level | |
| | | | Ranking or choice problems | DCV | Absolute | |
| | | | | | Rank related | |
| | | | | PWA | At the global level | Complete |
| | | | | | | Partial |
| | | | | | At the local level | Complete |
| | | | | | | Partial |
| | | | | CRIAO | At the global level | Complete |
| | | | | | | Partial |
| | | | | | At the local level | Complete |
| | | | | | | Partial |
| | | | | CRIAR | At the global level | |

| | | | | | |
|---|---|---|---|---|---|
| | | | | | At the local level |
| | | | | RRR | At the global level |
| | | | | | At the local level |
| | c.3.2 - Frequency of preference input | One time | | | |
| | | Incremental | | | |
| | c.3.3 - Confidence level | Accepted | | | |
| | | Not accepted | | | |
| **c.4 - Exploitation of the preference model** | c.4.1 - Univocal recommendation | c.4.1.1 - Without output variability analysis | Single model | Deterministic | Only 1 model, the only one available |
| | | | | Representative | Algorithmic |
| | | | | | Direct involvement |
| | | | Multiple models | | |
| | | c.4.1.2 - With output variability analysis | Compromise | | |
| | | | Representative | | |
| | c.4.2 - Output variability analysis | c.4.2.1 - How to conduct the output variability analysis | Extreme results with all compatible models -> All the models are taken - all the space is covered with mathematical programming (Necessary/possible) | | |
| | | | Distribution of results with a sample of | | |

| | | | | |
|---|---|---|---|---|
| | | | compatible models - A sample of all models considered with Monte Carlo simulations (Stochastic) | |
| | | c.4.2.2 - Results/outcomes that are the focus of the output variability analysis | Choice and ranking | Selection |
| | | | | Rank |
| | | | | Score |
| | | | | Preference relation |
| | | | | Preference intensity |
| | | | Sorting | Class assignment |
| | | | | Assignment-based pairwise relation |
| | | | | Class cardinality |

| AAC | APC | CRAIO | CRAIR | DCV | DCVC | FU | INT |
|---|---|---|---|---|---|---|---|
| Assignment of reference alternatives to categories | Assignment-based pairwise comparisons of alternatives | Comparisons of reference alternatives with respect to intensity of preference expressed on an ordinal scale | Comparisons of reference alternatives with respect to intensity of preference expressed on a ratio scale | Desired comprehensive values of alternatives | Desired comprehensive values of alternatives assigned to a given class, class range | Fuzzy | Intervals |
| ORAC | ORCWP | ORSC | PWA | PWCD | VAC | RRR | |
| Ordering all criteria | Ordering all criteria with intensities of preferences | Ordering some criteria | Pairwise comparisons of reference alternatives | Pairwise comparison of the differences between importance of criteria | Valued assignment of alternatives at the comprehensive level with different credibility degrees | Rank-related requirement | |

***Table 2:*** *Example of MCDA methods in the MCDA-MSS database, mapped according to a subset of features of the taxonomy for each type of problem statement. Bold "1" are those that are used to trigger the rules in the main text.*

| **MCDA methods** | | **Section 1 - Problem typology** | | | | | | | | | | | | | | | | | | **Section 2 - …** |
|---|---|---|---|---|---|---|---|---|---|---|---|---|---|---|---|---|---|---|---|---|
| | | c.1.1 - Problem statement | | | | | | | | | | | | | | | | | Other features | Other features |
| | | Ranking | | | | Sorting | | | | | | *Clustering* | | | | | Choice | | … | … |
| | | *Order* | | *Scale leading the recommendation* | | *Order of decision classes* | | *Scale leading the recommendation* | | *Cardinality* | | *Non-relational* | *Relational: Partial tournament* | *Relational: Complete tournament* | *Order* | | *Cardinality* | | | |
| | | *Partial* | *Complete* | *Ordinal* | *Cardinal* | *Partial* | *Complete* | *Ordinal* | *Cardinal* | *With constraints* | *Without constraints* | | | | *Partial* | *Complete* | *With constraints* | *Without constraints* | | |
| *For ranking problems* | | | | | | | | | | | | | | | | | | | | |
| 1 | AHP | 0 | 1 | 0 | 1 | 0 | 0 | 0 | 0 | 0 | 0 | 0 | 0 | 0 | 0 | 0 | 0 | 0 | … | … |
| 2 | GRIP | **1** | 0 | 0 | **1** | 0 | 0 | 0 | 0 | 0 | 0 | 0 | 0 | 0 | 0 | 0 | 0 | 0 | … | … |
| 3 | PROMETHEE II | 0 | 1 | 0 | 1 | 0 | 0 | 0 | 0 | 0 | 0 | 0 | 0 | 0 | 0 | 0 | 0 | 0 | … | … |
| *For sorting problems* | | | | | | | | | | | | | | | | | | | | |
| 4 | DRSA - sorting | 0 | 0 | 0 | 0 | 0 | 1 | 1 | 0 | 0 | 1 | 0 | 0 | 0 | 0 | 0 | 0 | 0 | … | … |
| 5 | ELECTRE SORT | 0 | 0 | 0 | 0 | 1 | 1 | 1 | 0 | 0 | 1 | 0 | 0 | 0 | 0 | 0 | 0 | 0 | … | … |
| 6 | ROR-UTADIS | 0 | 0 | 0 | 0 | 0 | **1** | 0 | **1** | **1** | 1 | 0 | 0 | 0 | 0 | 0 | 0 | 0 | … | … |
| *For clustering problems* | | | | | | | | | | | | | | | | | | | | |
| 7 | MCUC-CSA | 0 | 0 | 0 | 0 | 0 | 0 | 0 | 0 | 0 | 0 | 0 | 0 | 0 | 1 | 0 | 0 | 0 | … | … |
| 8 | P2CLUST | 0 | 0 | 0 | 0 | 0 | 0 | 0 | 0 | 0 | 0 | 0 | 0 | 0 | 0 | **1** | 0 | 0 | … | … |
| *For choice problems* | | | | | | | | | | | | | | | | | | | | |
| 9 | Constrained choice | 0 | 0 | 0 | 0 | 0 | 0 | 0 | 0 | 0 | 0 | 0 | 0 | 0 | 0 | 0 | 1 | 0 | … | … |
| 10 | ELECTRE I | 0 | 0 | 0 | 0 | 0 | 0 | 0 | 0 | 0 | 0 | 0 | 0 | 0 | 0 | 0 | 0 | **1** | … | … |
| | | | | | | | | | | | | | | | | | | | | |
| … | Method … | … | … | … | … | … | … | … | … | … | … | … | … | … | … | … | … | … | … | … |

***Table 3:*** *Example questions used to describe the MCDA methods and a case study (Adapted from Cinelli et al. (2020a))*

| Section | Question context | Questions used to describe the MCDA method | Questions used to describe the case study |
|---|---|---|---|
| 1: Problem typology | Problem statement | What type of decision recommendation does the method provide? | What type of decision recommendation is requested? |
| | Order of alternatives, if ranking as a problem statement was chosen | Does the method provide a partial or complete order of alternatives as a final decision recommendation? | Does the DM require a complete order of alternatives as a final decision recommendation, or a partial one would be enough? |
| | Structure of the set of criteria | Can the method accept a flat and/or hierarchical structure of the criteria? | Is the structure of the criteria flat or hierarchical? |
| 2: Preference model | Comparison of the performances on all the criteria | How does the method perform the comparison of the performances on the criteria? | How should the comparison of the performances on the criteria be performed? |
| | Compensation between criteria | What is the level of compensation between the criteria performances that the method implements? | How much can the good performance on a criterion compensate for the bad performance on another criterion? |
| 3: Elicitation of preferences | Type of weights, if the user wants criteria weights to be used | Does the method accept precise and/or imprecise weights? | Should precise or imprecise weights be used in this case study? |
| | Type of pairwise comparison thresholds if the user wants pairwise comparison thresholds to be used | Which type, if any, of pairwise comparison thresholds does the method accept? | What type of pairwise comparison thresholds should be used? |
| | Interactions between criteria | What type, if any, of interactions can the method handle? | What type, if any, of interactions should be handled? |
| 4: Exploitation of the preference model | Type of exploitation of the preference model | Which type of exploitation of the preference model does the method support? | What type of exploitation of the preference model should be applied for this case study? |
| | Type of output variability analysis, if the user indicated that output variability analysis be performed | What type, if any, of output variability analysis can the method perform? | How should the output variability analysis be conducted? |

***Table 4:*** *Example of a description of MCDA methods and case studies according to a sample of features of the taxonomy. ✓in "Description of the MCDA methods" = the method supports this feature; ✓in "Description of the case studies" = the case study requests this feature. The last three columns show the method(s) chosen by the authors of the case study, the recommendation of the MCDA-MSS with the relevant method(s) for each case study with complete match, and the one(s) with the missed features, respectively*
*(Adapted from Cinelli et al. (2020a) and Wątróbski et al. (2019))*

| | **c.1 - Problem typology** | | | | | | | **Other features** | **Method(s) chosen in the case study** | **MCDA method(s) recommended by the MCDA-MSS, complete match** | **MCDA method(s) recommended by the MCDA-MSS, with missed features** |
|---|---|---|---|---|---|---|---|---|---|---|---|
| | c.1.1 - Problem statement | | | c.1.2 – Set of alternatives | | c.1.3 – Criteria structure | | … | | | |
| | Ranking | Sorting | … | Stable | Incre-mental | Flat | Hierar-chical | | | | |
| **Description of the MCDA methods** | | | | | | | | | | | |
| Method 1 | | ✓ | | ✓ | ✓ | ✓ | | | | | |
| Method 2 | ✓ | | | ✓ | | ✓ | ✓ | | | | |
| Method … | | | ✓ | | ✓ | | ✓ | ✓ | | | |
| | | | | | | | | | | | |
| **Description of the case studies** | | | | | | | | | | | |
| Literature case study 1 | | ✓ | | | ✓ | ✓ | | | Method 1 | Method 1 | / |
| Literature Case study 2 | ✓ | | | ✓ | | ✓ | | ✓ | Method 2 | / | Method 3 & 4 |

***Table 5:*** *Summary of results of the MCDA-MSS test with the literature case studies*

| Case study | Reference | MCDA method chosen in the case study | Missed features between the MCDA method chosen in the case study and the case study description by the authors of this paper | MCDA method(s) recommended by the MCDA-MSS with complete match of features | Some of the MCDA method(s) recommended by the MCDA-MSS with fewest missed features shown in '[]' brackets |
|---|---|---|---|---|---|
| 1 | (Ferretti and Degioanni 2017) | MAVT | None | MAVT | / |
| 2 | (Dias et al. 2016) | SMAA-WAM | None | SMAA-WAM, HSMAA | / |
| 3 | (Dias et al. 2016) | VIP | None | VIP | / |
| 4 | (Liou 2011) | DRSA Variable Consistency (VC) - sorting | None | DRSA Variable Consistency (VC) - sorting | / |
| 5 | (Haurant et al. 2011) | ELECTRE IS | None | ELECTRE I, MCHP-ELECTRE, RUBIS | / |
| 6 | (Sparrevik et al. 2012) | SMAA-PROMETHEE | None | SMAA-PROMETHEE | / |
| 7 | (Maxim 2014) | WAM with performances transformation | 1. Use of number-coded qualitative-ordinal criteria<br>2. Use of weights meaning importance coefficients | / | QUALIFLEX [works with pairwise comparisons between the alternatives] |
| 8 | (Syarah Raudhatul and Firman 2019) | PROMETHEE II | 1. Use of AHP for weighting combined with other methods<br>2. No consideration of interactions among criteria | / | GAIA-SMAA-PROMETHEE-INT [criteria weights are used as importance coefficients and output variability analysis is the exploitation of the preference model] |
| 9 | (Qiu et al. 2017) | WAM with performances transformation | 1. Use of number-coded qualitative-ordinal criteria<br>2. Use of AHP for weighting combined with other methods<br>3. Problem statement sorting | / | FlowSort, PROMSORT [criteria weights are used as importance coefficients] |
| 10 | (El Mazouri et al. 2018) | ELECTRE III | 1. Problem statement sorting<br>2. Incremental set of alternatives<br>3. Incomplete set of criteria | / | ELECTRE SORT, OSMPOC, THESEUS [the set of criteria should be complete] |

# Electronic Supplementary Information (ESI)

*for the paper*

## Recommending Multiple Criteria Decision Analysis Methods with A New Taxonomy-based Decision Support System

Marco Cinelli[1,^,*], Miłosz Kadziński[1], Grzegorz Miebs[1], Michael Gonzalez[2], Roman Słowiński[1,3]

*[1] Institute of Computing Science, Poznan University of Technology, Piotrowo 2, 60-965 Poznań, Poland (marco.cinelli@put.poznan.pl; milosz.kadzinski@cs.put.poznan.pl; gregorz.miebs@cs.put.poznan.pl; roman.slowinski@cs.put.poznan.pl)*

*[2] Environmental Decision Analytics Branch, Land Remediation and Technology Division, Center for Environmental Solutions and Emergency Response, Office of Research and Development, U.S. Environmental Protection Agency, 26 West Martin Luther King Dr., Cincinnati, 45268, Ohio, USA (gonzalez.michael@epa.gov)*

*[3] Systems Research Institute, Polish Academy of Sciences, Newelska 6, 01-447 Warsaw, Poland*

*[*] Corresponding author: marco.cinelli@put.poznan.pl*

## Table of Contents

# Appendix A – Methods included in the MCDA-MSS database

This table provides the list of the MCDA methods in MCDA-MSS together with a brief description and their respective reference.

***Table S1:*** *Brief description of the MCDA methods included in the DSS*

| N. | Method name | Brief description | Reference |
|---|---|---|---|
| 1 | ACUTA | An extension of the UTA method to reformulate the development of the model selection procedure. | (Bous et al. 2010) |
| 2 | AHP | Alternatives are compared on a pairwise comparison basis with a pre-defined scale (the original is 0-9). The better the performance of an alternative with respect to the other, the higher is the score. Criteria weights are obtained in a similar manner, by comparing the criteria pairwise. After these comparisons are completed, the values in the performances and weighting matrices are combined (additively in the original AHP), to derive a final score for each alternative, which can be used to rank them. | (Saaty 1990) |
| 3 | AHP interval | It uses the same methodology to compare alternatives and priorities of criteria as AHP, but allows to use intervals instead of single numbers when performing the comparisons. | (Ahn 2017) |
| 4 | AHP Sort - Boundary profiles | A variant of AHP used for the sorting of alternatives into predefined ordered categories, using boundary profiles to delimit categories. | (Ishizaka et al. 2012) |
| 5 | AHP Sort - Characteristic profiles | A variant of AHP used for the sorting alternatives into predefined ordered categories, using reference alternatives to delimit categories | (Ishizaka et al. 2012) |
| 6 | ANP | The ANP method is a generalization of the AHP and its structure is a network rather than a hierarchy. Pairwise comparison matrices are used to characterize the interactions between the criteria and to then build a supermatrix. The latter is used to derive the global scores (priorities) of the alternatives. | (Saaty 2016) |
| 7 | ARAS | A complete ranking method using a linear normalization based on the sum of performances of all alternatives, aggregated by an additive weighted sum. | (Zavadskas and Turskis 2010) |
| 8 | ARAS-F | Extension of the ARAS method to handle fuzzy performances and preferences. | (Turskis and Zavadskas 2010) |
| 9 | ARGUS | It uses only the qualitative nature of the data. It transforms all the performances on a 5-point scale, from indifference to very strong preference. The weights are also expressed on a five-point scale preference. The overall preference model is developed based on an outranking relation, allowing for incomparability too. | (De Keyser and Peeters 1994) |
| 10 | ARIADNE | A ranking method that allows the use of imprecise criteria performances and preferences, resulting in a robustness-based exploitation of the preference model. | (Sage and White 1984) |
| 11 | ASSESS (MAUT) | A method similar to MAVT, where the normalization of the performances is conducted by means of utility instead of value functions. This modelling feature allows accounting for the risk attitude of the DM, which is not supported in MAVT. | (Farquhar 1984) |
| 12 | BWM | The Best Worst Method is similar to AHP for the weights elicitation. It uses target normalization for normalization of performance matrix if performance scales are not homogeneous. | (Rezaei 2015) |
| 13 | CBRM-I | Case-based distance model for sorting with cardinal data using preferences inferred from a set of assignment examples from the DM. | (Chen et al. 2007) |
| 14 | CBRM-II | Extension of CBRM-I to handle ordinal criteria, in addition to cardinal ones. | (Chen et al. 2007) |
| 15 | Choquet bipolar integral | It implements a similar modelling approach as the Choquet integral, with the added feature that it can account for antagonistic interactions between criteria. | (Greco and Rindone 2013) |
| 16 | Choquet integral | A method that builds a score for each alternative, by allowing the inclusion of interactions (positive and negative) between criteria. It can be seen as a generalized version of the weighted sum, which is a Choquet algorithm in cases where no interactions are present. Criteria measurement scales have to be normalized to before the aggregation step. | (Grabisch 1996) |
| 17 | CLUST-MCDA | A clustering method that supports problems with partially and completely ordered categories. The alternatives that are indifferent are grouped into clusters, which are then ranked, based on the preference | (Meyer and Olteanu 2013) |
| | | of the DM. Incomparability can be included too, so that partial instead of complete rankings can be obtained. | |
| 18 | COMET | Triangular fuzzy numbers are used to compare the performance of the alternatives with characteristic objects (CO) (i.e., fictitious alternatives). Based on these comparisons, activated fuzzy rules are summed to define the preference of each alternative, which is then used to rank them in a complete order. | (Sałabun 2015) |
| 19 | Compromise programming | A scoring method that implements a normalization approach based on the ideal and anti-ideal values in the set of alternatives. The final score is provided by an average of the weights of the criteria multiplied by the normalized criteria values. | (Romero and Rehman 2003) |
| 20 | Constrained choice | A method to support choice problems with a constrained set of alternatives. It operated on binary relations, exploiting the ordinal nature of the input data. | (Podinovski 2010) |
| 21 | COPRAS | A ranking method using a target-based normalization of the criteria. Weighted normalized values for each criterion are calculated and used to derive maximizing and minimizing indices, according to whether the criteria should be maximized or minimized. A function called relative significance is then used to derive the utility of each alternative and provide the ranking. | (Zavadskas et al. 2004) |
| 22 | CORT | A preference disaggregation method exploiting assignment examples by the DM, assuming that the marginal value functions are dependent on the performances of the alternatives. | (Kadziński et al. 2020a) |
| 23 | CUT | A method for choice and ranking problems using intensities of preference as the main indirect preference information. | (Argyris et al. 2014) |
| 24 | DEX | A qualitative method that integrates the multiple criteria modeling with a rules-based approach. The decision rules must be defined for all the criteria in the model and they can be either directly defined by the DM or inferred from holistic preferences. | (Bohanec et al. 2013) |
| 25 | DIS-CARD-OUTRANKING | A method for partial and complete multiple criteria sorting, exploiting an outranking-driven model. It accepts DM's preferences in the form of exemplary assignments of some reference alternatives and also desired cardinalities of classes. | (Kadziński and Słowiński 2013) |
| 26 | DIS-CARD-UTADIS | A method for complete multiple criteria sorting, accepting the same DM's preferences as DIS-CARD – outranking-driven sorting, but exploiting an additive value function-driven model. | (Kadziński and Słowiński 2013) |
| 27 | DISWOTH | A distance-based sorting method that does not require the DM to define class profiles. | (Karasakal and Civelek 2020) |
| 28 | DRSA - ranking | Adaptation of DRSA to handle ranking problems. The method works with a set complete or partial pairwise comparisons of reference alternatives. | (Szeląg et al. 2013) |
| 29 | DRSA - sorting | Extension of classical rough set approach to preference-ordered input information (i.e., criteria) to handle sorting problems. It uses DM's assignments of reference alternatives to classes to derive a sorting model based on "if …, then …" decision rules.<br>It requires that alternatives having a not-worse evaluation with respect to the criteria cannot be assigned to a respective worse class. | (Greco et al. 2001a) |
| 30 | DRSA - sorting (new classification scheme) | It applies the same working procedure as DRSA sorting, except for the type of scale used to provide the recommendation, which is cardinal with this method. | (Błaszczyński et al. 2009) |
| 31 | DRSA – sorting: monotonicity discovery (new classification scheme) | Extension of DRSA for sorting that can include criteria whose monotonicity is not known and has to be discovered. It provides a cardinal recommendation like DRSA – sorting. | (Błaszczyński et al. 2012) |
| 32 | DRSA - sorting imprecise | Extension of DRSA sorting to account for imprecise performances on the criteria and assignments of the reference alternatives. | (Dembczyński et al. 2009) |
| 33 | DRSA VC - ranking | It is the same as DRSA ranking, with the added capacity of including inconsistencies in the assignments as preference information. | (Szeląg et al. 2013) |
| 34 | DRSA VC - sorting | Extension of classical DRSA sorting so that it can accept a limited proportion of inconsistent examples (i.e., alternatives). This means that some inconsistencies in the assignments (e.g., alternative A performs at least as well as alternative B, but it is assigned by the DM to a worse class) can be accepted as preference information.<br>The range of this relaxation in the dominance principle is controlled by an index called consistency level. | (Greco et al. 2001b) |
| 35 | DRSA VC - sorting (new classification scheme) | It applies the same working procedure as DRSA Variable Consistency (VC) - sorting, except for the type of scale used to provide the recommendation, which is cardinal with this method. | (Błaszczyński et al. 2009) |
| 36 | DRSA VC – sorting: monotonicity discovery (new classification scheme) | Extension of DRSA – sorting: monotonicity discovery, with the added capacity of including inconsistencies in the assignments as preference information. | (Błaszczyński et al. 2012) |
| 37 | Dual weak-strong CI | A composite indicator based on a data-drive normalization of the performances of the critera, which accommodates a weak and strong compensation capability | (Cabello et al. 2018) |
| 38 | EDAS | A ranking method which involves comparison of each alternative with average profiles computed automatically. Based on the distances from the positive and negative distance from the average, a score is derived for each alternative. | (Keshavarz Ghorabaee et al. 2015) |
| 39 | ELECTRE GKMS | A robust ordinal regression-based ranking method to construct a set of outranking models compatible with preference information in the form of pairwise comparison of reference alternatives. Such preference information can have varying degrees of confidence and the preference thresholds can be precise and imprecise. | (Greco et al. 2011a) |
| 40 | ELECTRE I | The first outranking method proposed by Bernard Roy, which operates on a pairwise comparison between alternatives. It allows to deal with choice problems where measurement scales should be numerical with identical ranges and no thresholds are accepted. | (Figueira et al. 2016) |
| 41 | ELECTRE II | One of the first outranking methods developed for ranking problems. It is a true criteria-based procedure, with the allowance of veto condition. | (Figueira et al. 2016) |
| 42 | ELECTRE III | Extension of ELECTRE II method to include pseudo-criteria, meaning that indifference and preference thresholds are now included in the preferences modelling. The ranking of the alternatives is driven by a set of concordance and discordance indices, defining how well and poorly each alternative performs with respect to each other. | (Figueira et al. 2016) |
| 43 | ELECTRE III-H | Extension of the ELECTRE III method to deal with hierarchically structured criteria. | (Del Vasto-Terrientes et al. 2015) |
| 44 | ELECTRE III-INT | Extension of the ELECTRE III method to deal with interacting criteria. | (Bottero et al. 2015) |
| 45 | ELECTRE IS | An extension of ELECTRE I (choice problems) to account for pseudo-criteria, so that indifference, preference and veto thresholds are added to the modelling of preferences. | (Figueira et al. 2016) |
| 46 | ELECTRE IV | A ranking method similar to ELECTRE III, which however does not require the use of weights. | (Figueira et al. 2016) |
| 47 | ELECTRE Iv | Extension of ELECTRE I (choice problems) to include the veto threshold and allow the use of heterogeneous measurement scales. | (Figueira et al. 2016) |
| 48 | ELECTRE SORT | An ELECTRE-based sorting method capable of dealing with incomparable classes while taking into account an unlimited set of criteria. The classes in these problems can be partially ordered, meaning that the classes have a natural ordering on each criterion but that this ordering is not the same for all the criteria. Therefore, classes can be incomparable. | (Ishizaka and Nemery 2014) |
| 49 | ELECTRE SORT-KMOB | A sorting method that does not require the definition of profiles and instead it uses assignment of reference alternatives to classes. | (Köksalan et al. 2009) |
| 50 | ELECTRE TRI-B | A sorting method that tackles problems with complete ordinal set of classes. It employs pseudo-criteria and the alternatives are compared to a set of class profiles to assess whether they are at least as good as them. These profiles are boundary conditions delimiting the performance to one class or another. | (Figueira et al. 2016) |
| 51 | ELECTRE TRI-B-disaggregation | A method based on indirect preferences elicitation (e.g., pairwise comparisons on the criteria, assignment of reference alternatives to categories) to elicit the operational parameters of ELECTRE-TRI-B. | (Dias et al. 2002) |
| 52 | ELECTRE TRI-B-H | Extension of the ELECTRE TRI-B method to deal with hierarchically structured criteria. | (Vasto-Terrientes et al. 2016) |
| 53 | ELECTRE TRI-C | A sorting method similar to ELECTRE TRI B where the distinctive feature is that each category is defined by a single characteristic reference alternative. | (Almeida-Dias et al. 2010) |
| 54 | ELECTRE TRI-nB | Extension of ELECTRE TRI B to account for multiple boundary profiles delimiting the preference ordered classes. | (Fernández et al. 2017) |
| 55 | ELECTRE TRI-nC | Extension of ELECTRE TRI C to account for multiple reference alternatives defining the preference ordered classes. | (Almeida-Dias et al. 2012) |
| 56 | ELECTRE TRI-rC | A modified version of ELECTRE TRI C using a different type of assignment rules. | (Kadziński et al. 2015c) |
| 57 | ERA | Extension of ROR for the analysis of extreme ranking results and for choice problems. | (Kadziński et al. 2012a) |
| 58 | EVAMIX | A ranking method that builds concordance indices tailored to the type of input data, either qualitative or quantitative. The resulting ranking is a combination of such indices with the inclusion of the weighting information on the criteria, which can be precise (quantitative) or imprecise (qualitative). | (Voogd 1982) |
| 59 | FITradeoof - choice | This method offers a simplified (compared to traditional trade-off method) weights elicitation technique for trade-offs to be applied in MAVT-driven choice problems. | (de Almeida et al. 2016) |
| 60 | FITradeoof – ranking | This method offers a simplified (compared to traditional trade-off method) weights elicitation technique for trade-offs to be applied in MAVT-driven ranking problems. | (Frej et al. 2019) |
| 61 | Flowsort | A sorting method to classify alternatives to completely ordered classes defined with limiting profiles or reference alternatives. The assignment of the alternatives depends on a global comparison with all the profiles simultaneously. The class assignment is based on the PROMETHEE I or PROMETHEE II ranking method. | (Nemery and Lamboray 2008) |
| 62 | FPP | A sorting method based on outranking relations accepting single and multiple boundary profiles. | (Perny 1998) |
| 63 | Fuzzy AHP | This method extends the AHP methodology by allowing the use of fuzzy numbers in the pairwise comparison between alternatives. This uncertainty can be related to factors external to the DM (e.g., measurement scale) or dependent on the DM, like due to information overload, stress, limited attention. | (Ayhan 2013) |
| 64 | Fuzzy ANP | An extension of the ANP method that allows to account for the pairwise comparisons with fuzzy numbers. This extension aims at including vagueness and uncertainty in the structuring of the problem and the DM's preferences. | (Promentilla et al. 2008) |
| 65 | Fuzzy MIN_MAX | A ranking method that accepts fuzzy performances of the alternatives and develops a decision recommendation based on the best and worst performances in the criteria set. | (Dubois et al. 1988) |
| 66 | Fuzzy MULTIMOORA | MULTIMOORA method with the extension of linguistic performances treated as fuzzy numbers | (Brauers et al. 2011) |
| 67 | Fuzzy PROMETHEE I | An extension of PROMETHEE I to include fuzzy numbers modelling to account for vagueness and uncertainty in the performance data and the DM's preferences. It provides a partial ranking of the alternatives. | (Geldermann et al. 2000) |
| 68 | Fuzzy PROMETHEE II | An extension of PROMETHEE II to include fuzzy numbers modelling to account for vagueness and uncertainty in the performance data and the DM's preferences. It provides a complete ranking of the alternatives. | (Geldermann et al. 2000) |
| 69 | Fuzzy TOPSIS | An extension of TOPSIS method to include fuzzy numbers modelling to account for vagueness and uncertainty in the performance data and the DM's preferences.<br>The distance between the alternative, the positive ideal solution (PIS) and negative ideal solution (NIS) is the sum of distances between fuzzy numbers representing each criterion separately. | (Nădăban et al. 2016) |
| 70 | Fuzzy VIKOR | An extension of VIKOR method to include fuzzy numbers modelling to account for vagueness and uncertainty in the performance data and the DM's preferences. | (Opricovic 2011) |
| 71 | Fuzzy weighted average | A fuzzy version of the weighted average method, where weights and criteria performances are presented by means of fuzzy numbers. | (Guh et al. 2008) |
| 72 | GAIA-SMAA-PROMETHEE-INT | An extension of PROMETHEE to account for hierarchy of interacting criteria and robustness of the recommendation with stochastic and robust ordinal regression. | (Arcidiacono et al. 2018) |
| 73 | GRIP | Extension of $UTA^{GMS}$ to include a new type of indirect preferences, i.e., comparisons of reference alternatives with respect to intensity of preference. | (Figueira et al. 2009) |
| 74 | HSMAA | Hierarchical Stochastic Multiobjective Acceptability Analysis, based on an additive weighted sum. | (De Matteis et al. 2019) |
| 75 | IDRA | A ranking method that allows including trade-offs and importance coefficients to develop a mixed utility function. This function is exploited to derive a preference index that defines the probability that one alternative is preferred to another. Each piece of intercriteria information, defined as a decision rule, is assigned a credibility value that is used to drive the definition of the global preference for each alternative. | (Greco 1997) |
| 76 | Imprecise ROR | A ROR-based ranking method that accepts imprecise performances of the alternatives. | (Corrente et al. 2017c) |
| 77 | INTERCLASS-B | A sorting method inspired by ELECTRE TRI-B that can handle uncertainty in criteria performances, weights, veto thresholds, cutting level, and limiting profiles. Classes are delimited by single boundary profiles. | (Fernández et al. 2019a) |
| 78 | INTERCLASS-nB | Extension of INTERCLASS-B to account for multiple boundary profiles to distinguish the classes. | (Fernández et al. 2019b) |
| 79 | INTERCLASS-nC | A sorting method inspired by ELECTRE TRI-nC that can handle uncertainty in criteria performances, weights, veto thresholds, cutting level, and reference alternatives. Classes can be delimited by multiple reference alternatives. | (Fernández et al. 2019b) |
| 80 | Lexicographic method | A simple ranking method that operates using a complete ranking of the criteria, without ties. The alternatives are compared pairwise, starting from the most important criterion. If alternative A is strictly better than another one (B) on the most important criterion, then A is declared globally preferred to B without even considering the other criteria. If A and B are indifferent on such criterion, the comparison moves to the next criterion, and so on. | (Bouyssou et al. 2006) |
| 81 | MACBETH | A ranking method using a qualitative seven-point scale to compare alternatives. These comparisons are used to develop quantitative value functions. An adapted SWING weighting methodology employing the seven-point scale allows to derive the weights as scaling constants. The additive weighted sum is used to aggregate the normalized and weighted criteria performances in an overall score. | (Bana E Costa and Vansnick 1999) |
| 82 | MACBETHSort | Adaptation of the MACBETH method for sorting problems. | (Ishizaka and Gordon 2017) |
| 83 | MAGIQ | A ranking method that uses rank order centroids to assign relative weights to the criteria and performances of the alternatives. | (McCaffrey 2009) |
| 84 | MAPPACC | A pairwise-comparison based method that normalizes the performances of the alternatives on a 0-1 scale and aggregates them with several relative preference indices, accounting for the weight of the criteria too. | (Matarazzo 1986) |
| 85 | MAVT | A ranking method that develops a single score for each alternative by aggregating the normalized criteria and their weights, which in this case are trade-offs. The normalized criteria values are obtained by means of value functions. | (Bottero et al. 2014) |
| 86 | MCHP-Choquet | Extension of the Choquet integral method to handle hierarchies of criteria. It is possible to compare two alternatives not only globally, but also partially, taking into account a particular subset of criteria and the possible interaction between them. | (Angilella et al. 2013) |
| 87 | MCHP - ELECTRE | Extension of ELECTRE methods for choice to the case of the hierarchy of criteria. | (Corrente et al. 2013) |
| 88 | MCHP-ELECTRE-TRI-INT | Extension of ELECTRE-TRI to handle hierarchies of criteria and interactions between them. | (Corrente et al. 2016) |
| 89 | MCHP-PROMETHEE | Extension of PROMETHEE methods for ranking to the case of the hierarchy of criteria. | (Corrente et al. 2013) |
| 90 | MCHP-SMAA-Choquet | A method that allows tackling problems with hierarchical structures of (possibly) interacting criteria, accounting also for the uncertainty, using a stochastic approach, for the criteria weighting as well as their interactions. | (Angilella et al. 2018) |
| 91 | MCHP-SMAA-ELECTRE-III-INT | Extension of ELECTRE III to handle hierarchies of criteria, interactions between them and exploitation of the preference model with stochastic robustness considerations. | (Corrente et al. 2017b) |
| 92 | MCHP-UTA-ROR | Combination of Multiple Criteria Hierarchy Process (MCHP) and ROR for ranking problems. This allows to account for preference information at different levels in the hierarchy and derive necessary and preference relations at each node in the hierarchy. | (Corrente et al. 2012) |
| 93 | MCHP-UTADIS | Extension of the UTADIS method for sorting problems to deal with hierarchies of criteria by means of the combination with the Multiple Criteria Hierarchy Process (MCHP). | (Corrente et al. 2017a) |
| 94 | MCHP-UTADIS-GMS | Extension of MCHP-UTADIS with ROR to account for the whole set of instances of the preference model compatible with the provided preference information. | (Corrente et al. 2017a) |
| 95 | MCUC-CSA | A method to cluster and sort alternatives considering multi-criteria categories with a partial order structure. The distinctive feature with respect to sorting methods is that at the problem formulation stage, the number of categories and their characteristics is unknown. The ranking of the categories follows an outranking-based strategy. | (Rocha et al. 2013) |
| 96 | MELCHIOR | Pseudo-criteria with indifference and preferences thresholds are used to build concordance analysis following the outranking paradigm, which similarly to ELECTRE IV leads to a ranking of the alternatives, without the use of any weighting. | (Vincke 1999) |
| 97 | MHDIS | A sorting method based on a hierarchical assignment of the alternatives, starting from the best group down to the worst one. | (Zopounidis and Doumpos 2000a) |
| 98 | MOORA | A scoring method based on a ratio-driven normalization. It does not use weights and it aggregates the normalized performances accounting for the distance from the reference (i.e., best) performance in the set of alternatives. | (Brauers and Zavadskas 2006) |
| 99 | MPOC | A method for partial clustering based on a crisp outranking relation, exploiting the preferences of the DM in the form of pairwise comparison thresholds and weights. | (Rocha and Dias 2013) |
| 100 | MR-Sort | An adaptation of the ELECTRE TRI-B sorting model without the use of pairwise comparison thresholds. | (Bouyssou and Marchant 2007) |
| 101 | MR-Sort-Card | Adaptation of the MR-Sort method to deal with sorting problems where constrains can be defined to set the number of alternatives to be assigned to each class. | (Özpeynirci et al. 2018) |
| 102 | MR-Sort-Imprecise | Extension of the MR-Sort method to handle problems with imprecise or missing performances. | (Meyer and Olteanu 2019) |
| 103 | MULTIMOORA | Adaptation of the MOORA method by using the multiplicative operator to develop the final score. | (Brauers and Zavadskas 2010) |
| 104 | N-Tomic | A sorting method based on exploitation of the outranking relations, allowing different degrees of compensation. | (Massaglia and Ostanello 1989) |
| 105 | NAIADE | A partial and complete ranking method that uses fuzzy pseudo-criteria (i.e., including indifference and preference thresholds) to compare alternatives. Six preference relations are represented with fuzzy numbers and the information on the pairwise performance of the alternatives is used to aggregate these evaluations in order to take all the criteria into account at the same time. The final ranking is obtained by exploiting the positive (leaving) and negative (entering) flows, like in PROMETHEE. | (Munda 1995) |
| 106 | NAROR-Choquet | Adaptation of the non-additive robust ordinal regression to the Choquet integral, allowing the exploration of all possible and necessary preference models. | (Angilella et al. 2010) |
| 107 | NAROR-HC-Sort | A sorting method that handles hierarchically structured criteria with possible interactions between them. ROR and SMAA are also integrated in order to assess the robustness of the recommendations. | (Arcidiacono et al. 2021) |
| 108 | NFS-ROR-SOR | An ordinal and cardinal ranking method based on robust and stochastic ordinal regression. | (Kadziński and Michalski 2016) |
| 109 | Non compensatory algorithm | A method that produces a complete ordinal ranking, exploiting only the ordinal character of the data and using weights as importance coefficients. The ranking is driven by the pairwise comparison of alternatives according to the whole set of criteria. | (Munda and Nardo 2009) |
| 110 | Non-monotonic UTA | Extension of the UTA method to deal with non-monotonic criteria. | (Despotis and Zopounidis 1995) |
| 111 | Non-monotonic UTADIS-EO | Extension of the UTADIS method to deal with non-monotonic criteria. | (Doumpos 2012) |
| 112 | ORCLAS | An ordinal classification method that works according to decision rules defined directly in interaction with the DM. These rules are defined according to the sorting provided by the DM on the most informative alternatives, being the ones that allow to infer other rules. | (Larichev and Moshkovich 1997a) |
| 113 | ORESTE | The ordinal information on the position of the alternatives and the importance ranking of the criteria is used as input. Preference, indifference and incomparability relations are used to derive intermediate rankings, which are then aggregated to derive a final ranking. | (Pastijn and Leysen 1989) |
| 114 | OSMPOC | An outranking-based sorting method for partially ordered categories | (Nemery 2008) |
| 115 | OWA | Ordered weighted average. It is an aggregation function of the same type as the weighted sum, but the distinctive feature is that an additional set of weights can be assigned according to the relative position of the performances in the set. | (Yager 1993) |
| 116 | P2CLUST | A complete clustering method based on the PROMETHEE II method. | (Smet 2013) |
| 117 | PAIRCLASS | PROMETHEE-based sorting method based on indirect preference information in the form of reference alternatives assigned to classes. | (Doumpos and Zopounidis 2004) |
| 118 | PAIRS | A value functions-driven ranking method that accepts imprecise preferences in the form of interval judgements and pairwise comparisons, as well as interval performances of the alternatives. | (Salo and Hämäläinen 1992) |
| 119 | PAMSSEM I | A combination of key features of ELECTRE (for construction of the outranking relation), PROMETHEE (for exploitation of the relations), and NAIADE (for handling fuzzy evaluations) methods. It can deal with crisp, heterogenous, missing and uncertain data. It uses pseudo-criteria with the possibility of also veto thresholds. It computes a concordance, local discordance and outranking indices as in ELECTRE. The outranking degree is then obtained from the concordance matrix, similarly to the procedure used in PROMETHEE. Incoming and outgoing preference flows are lastly calculated and a partial ranking is derived. | (Alinezhad and Khalili 2019a) |
| 120 | PAMSSEM II | The same procedure as in PAMSSEM I is used. The difference is that a final complete ranking is obtained from the net outranking flow, derived from the difference of the positive and negative flows. | (Alinezhad and Khalili 2019a) |
| 121 | PASA | An outranking-based sorting method using categories defined by a set of alternatives sorted by the DM. The method constrains the categories where the alternatives can be assigned based on the previous choices made by the DM. | (Rocha and Dias 2008) |
| 122 | PCLUST | An extension of PROMETHEE I to obtain individual and interval clusters. It uses the FLOWSORT method for assignment of alternatives. PROMETHEE I is used on the set of central profiles and each alternative's performance. The resulting ranking is used to determine the cluster of each alternative. | (Sarrazin et al. 2018) |
| 123 | PDTOPSIS-Sort | A sorting method that uses preference disaggregation approached to infer the TOPSIS-sort parameters. | (de Lima Silva et al. 2020) |
| 124 | PLVF | A scoring method based on different types of piecewise linear value functions. | (Rezaei 2018) |
| 125 | Preference programming | An extension of the AHP method to accept imprecise ratio comparisons in the form interval ranges. | (Salo and Hämäläinen 1995) |
| 126 | Preference programming IOI | A value functions-based method that accepts incomplete ordinal information about the performances of the alternatives and the criteria weights. | (Punkka and Salo 2013) |
| 127 | PRIME | A generalisation of the PAIRS method, with the added capability of handling holistic judgments. | (Salo and Hamalainen 2001) |
| 128 | PROMETHEE CLUSTER | A clustering method for non-ordered clusters exploiting the flow-based procedure implemented in PROMETHEE. | (Cailloux et al. 2007) |
| 129 | PROMETHEE GKS | Robust ordinal regression applied to PROMETHEE I and II. | (Kadziński et al. 2012a) |
| 130 | PROMETHEE I | A partial ranking method that exploits pairwise comparisons, which are driven by differences of performances between the alternatives. Weights of criteria are used as importance coefficients and preference and indifference thresholds can be included to account for uncertainty in the performances as well as hesitation of the DM. PROMETHEE ranking is driven by two flows, a positive and a negative one. The first characterizes how much an alternative outranks the others, while the latter defines the opposite information. The intersection of these flows leads to the partial ranking in PROMETHEE I. | (Brans and De Smet 2016) |
| 131 | PROMETHEE II | It applies the same methodology as PROMETHEE I, with the difference that it provides a complete ranking of the alternative determined by the difference between the leaving and entering flows. | (Brans and De Smet 2016) |
| 132 | PROMETHE TRI | A sorting method based on the PROMETHEE methodology, exploiting characteristic profiles instead of boundary profiles (like in ELECTRE TRI) to sort the alternatives. | (Figueira et al. 2005) |
| 133 | PROMETHEE V | It extends PROMETHEE II by accepting modelling constraints that provide a subset of alternatives that best match the requirements from the DM. | (Brans and Mareschal 1992) |
| 134 | PROMSORT | A sorting method based on the PROMETHEE methodology, using boundary profiles to drive the classification. | (Araz and Ozkarahan 2007) |
| 135 | QUALIFLEX | This method is based on the evaluation of all possible rankings (permutations) of alternatives. The final ranking is determined based on the ranking of the alternatives with respect to each criterion. For each permutation of alternatives, a concordance/discordance index is developed, reflecting the concordance and discordance of their ranks. The final ranking is driven by the permutation whose ranking best reflects the preorders with respect to the whole set of criteria. | (Paelinck 1976) |
| 136 | REGIME | It provides a ranking of the alternatives based on concordance analysis, accounting for the set of criteria where one alternative is at least as good as another. A regime matrix with the pairwise comparisons of concordance for the alternatives is exploited to search for a consistent ranking of the alternatives. | (Hinloopen et al. 1983) |
| 137 | REMBRANT | The method uses pairwise comparisons of the decision makers with respect to the preference (or not) of one alternative with respect to the other, similarly to the AHP method. The geometric function is then used to aggregate the gradation levels expressed from the pairwise comparisons. | (Van den Honert and Lootsma 2000) |
| 138 | RICH | A ranking method that allows to specify subsets of criteria that contain the most important one, or to define multiple rankings with a certain set of criteria. | (Salo and Punkka 2005) |
| 139 | RMP-PD | A ranking method that develops the decision recommendation based on a list of pairwise comparisons provided by the DM. These preferences are used to elicit the weights, reference profiles and the lexicographic order on reference profiles. | (Olteanu et al. 2021) |
| 140 | ROR-Distance | Application of ROR to distance-based functions used to rank a set of alternatives using non-complete pairwise comparisons (also with intensities of preference) of reference alternatives. | (Zielniewicz 2017) |
| 141 | ROR-DRSA-Rank | Integration of ROR with DRSA for ranking to develop robust relations using all compatible sets of decision rules. | (Kadziński et al. 2015b) |
| 142 | ROR-DRSA-Sort | Integration of ROR with DRSA for sorting to develop robust relations using all compatible sets of decision rules. | (Kadziński et al. 2014) |
| 143 | ROR-ELECTRE-SORT | An ELECTRE-based multiple criteria sorting method that accounts for robustness of decision recommendations. Compared to competitive methods, it accepts desired class cardinalities. | (Kadziński and Ciomek 2016) |
| 144 | ROR-ELECTRE-TRI-B | Extension of ELECTRE TRI-B with ROR by allowing the provision of assignment examples, assignment-based pairwise comparisons, and desired class cardinalities. | (Kadziński and Martyn 2020) |
| 145 | ROR-ELECTRE-TRI-rC | Combination of ROR and ELECTRE TRI-rC to exploit the whole set of compatible models. | (Kadziński et al. 2015c) |
| 146 | ROR-PROMETHEE-SORT | A PROMETHEE-based multiple criteria sorting method that accounts for robustness of decision recommendations. Compared to competitive methods, it accepts desired class cardinalities. | (Kadziński and Ciomek 2016) |
| 147 | ROR-UTADIS | Extension of UTADIS to included ROR and also desired class cardinalities | (Kadziński et al. 2015a) |
| 148 | RUBIS | A method using outranking relations to deal with choice problems, including missing performances. | (Bisdorff et al. 2008) |
| 149 | RUTA | A preference disaggregation method for ranking purposes that infers additive value functions derived from the desired ranks and/or scores of some reference alternatives. | (Kadziński et al. 2013) |
| 150 | S-RMP | A ranking method based on external reference profiles using an outranking modelling. | (Rolland 2013) |
| 151 | SIR-TOPSIS | A partial and complete raking method that, similarly to TOPSIS, builds the recommendation based on the distance from the ideal and anti-ideal performances. | (Xu 2001) |
| 152 | SMAA | A method developed to deal with choice problems characterized by uncertain criteria performances and/or preferences. | (Lahdelma et al. 1998) |
| 153 | SMAA III | Extension of ELECTRE III to deal with imprecise performances of alternatives and preferences. | (Tervonen et al. 2008) |
| 154 | SMAA-2 | A ranking method that handles stochastically uncertain criteria performances and/or preferences. | (Lahdelma and Salminen 2001) |
| 155 | SMAA-AHP | Extension of the AHP method to deal with uncertain pairwise comparisons using a stochastic modelling. | (Durbach et al. 2014) |
| 156 | SMAA-based-RANK | A complete and ordinal ranking method that exploits a stochastic modelling for criteria performances preferences. It also provide a compromise univocal recommendation. | (Vetschera 2017) |
| 157 | SMAA-Choquet | Application of Stochastic Multicriteria Acceptability Analysis (SMAA) to the Choquet integral to account for uncertain performances and preferences in interaction-based aggregation | (Angilella et al. 2015) |
| 158 | SMAA-ELECTRE-I | Extension of the ELECTRE I method to handle stochastically uncertain criteria performances and/or preferences. | (Govindan et al. 2019) |
| 159 | SMAA-O | A choice and ranking method tailored to problems with at least one ordinal criterion, using stochastically uncertain criteria performances and/or preferences. | (Lahdelma et al. 2003) |
| 160 | SMAA-PROMETHEE | An extension of the PROMETHEE I and II ranking method with SMAA, which allows to account for uncertain performances of the criteria and preferences | (Corrente et al. 2014) |
| 161 | SMAA-TRI | An extension of the ELECTRE TRI sorting mehtod with SMAA, which allows to account for uncertain performances of the criteria and preferences | (Tervonen 2014) |
| 162 | SMAA-WAM | SMAA applied to weighted additive mean to rank alternatives by using a stochastic exploitation of the preference model. | (Greco et al. 2018) |
| 163 | SMAA- WGM | SMAA applied to weighted geometric mean to rank alternatives by using a stochastic exploitation of the preference model. | (Greco et al. 2018) |
| 164 | SMARTS | A ranking method based on normalization of the performances of the alternatives according to an arbitrary value function. The weights are calculated based on the SWING method. The final score of the alternatives is the weighted sum of the criteria weights and the normalized performances. | (Edwards and Barron 1994) |
| 165 | SOR-Rank | A ranking method that combines ROR and SMAA modelling, using indirect preference information in the form of pairwise comparisons of reference alternatives. | (Kadziński and Tervonen 2013a) |
| 166 | SOR-Sort | A sorting method that combines ROR and SMAA modelling, using indirect preference information in the form of assignment of reference alternatives to classes. | (Kadziński and Tervonen 2013b) |
| 167 | Sugeno integral | A ranking method tailored to qualitative interacting criteria, providing a score for each alternative. | (Grabisch and Labreuche 2016) |
| 168 | TACTIC | The modelling allows to include true-criteria and quasi-criteria, with indifference and veto thresholds. The procedure for the recommendation of the chosen alternative is similar to the one used in ELECTRE I, though in TACTIC the preference relation leads the development of the recommendation, while ELECTRE I uses the outranking relation. | (Vansnick 1986) |
| 169 | THESUS | A sorting method using fuzzy outranking relations defined according to the whole set of alternatives. | (Fernandez and Navarro 2011) |
| 170 | TODIM | It is a pairwise comparison based ranking method that does not use any thresholds and exploits the differences between the performances of the alternatives. | (Alinezhad and Khalili 2019b) |
| 171 | TOMASO | A sorting method based pairwise comparisons of alternatives, aggregated by means of the Choquet integral, whose fuzzy measures can be learnt from a set of reference alternatives assigned to classes by the DM. | (Marichal et al. 2005) |
| 172 | TOPSIS | A method used to rank alternatives based on the minimization of the distance from an ideal point and maximization of distance from a non-ideal one. The ideal point is a composite of the best performances on each criterion for every alternative. The anti-ideal point is the opposite and considers the worst performances of the alternatives. | (Bilbao-Terol et al. 2014) |
| 173 | TOPSIS interval | An extension of the TOPSIS method to include interval performances of the alternatives. | (Jahanshahloo et al. 2006) |
| 174 | TOPSIS interval and fuzzy | Extension of the TOPSIS method to include fuzzy and interval performances of the alternatives, as well as preferences of the DM. | (Chen and Tsao 2008) |
| 175 | TOPSIS Sort-B | An adaptation of the TOPSIS method for sorting problems using boundary profiles. | (de Lima Silva and de Almeida Filho 2020) |
| 176 | TOPSIS-Sort-C | An adaptation of the TOPSIS method for sorting problems using characteristic profiles. | (de Lima Silva and de Almeida Filho 2020) |
| 177 | UTA | Piecewise linear functions for each criterion are derived from holistic rankings of reference alternatives. One value function per criterion is then used to define the overall performance of new alternatives and to rank them accordingly. | (Jacquet-Lagreze and Siskos 1982) |
| 178 | UTA-GMS | This method applies robust ordinal regression to elicit additive value functions to then combine them in a single score to rank the alternatives. All the additive value functions are compatible with the indirect preference information are accepted. Instead of UTA, the value functions can be of the general form and the DM can provide a non-complete preference on the reference alternatives. | (Greco et al. 2008) |
| 179 | UTA-GMS-INT | An extension of the $UTA^{GMS}$ to include ‘‘bonus’’ or ‘‘penalty’’ values for positively or negatively interacting pairs of criteria, respectively. | (Greco et al. 2014) |
| 180 | UTA-NM | Extension of the UTA method to deal with non-monotonic criteria. | (Kliegr 2009) |
| 181 | UTA-NM-GRA | Another extension of the UTA method to deal with non-monotonic criteria. | (Ghaderi et al. 2017) |
| 182 | UTA-PLUS | Extension of the UTA method to include intensity of preferences and direct involvement in the choice of the representative single model for the development of the univocal recommendation. | (Kostkowski and Słowiński 1996) |
| 183 | UTA-poly-splines | An extension of the UTA method to consider non-linear marginals that can more accurately represent the DM's preferences. | (Sobrie et al. 2018) |
| 184 | UTA-ROBUST-REPRESENT – | Extension of UTA GMS and GRIP for ranking and choice problems, to derive a unique representative function from the results of the robust ordinal regression. | (Kadziński et al. 2012b) |
| 185 | UTADIS | UTA method adapted for sorting problems. | (Zopounidis and Doumpos 2000b) |
| 186 | UTADIS-ER | A sorting method based on an evidential reasoning approach using a set of assignment examples as main preference information. | (Liu et al. 2015) |
| 187 | UTADIS-GMS | UTA GMS adapted for sorting problems. | (Greco et al. 2010) |
| 188 | UTADIS-INT | Extension of the UTADIS method to deal with interacting criteria. | (Liu et al. 2020a) |
| 189 | UTADIS-KB | Extension of the UTADIS method to reduce misclassification problems. | (Köksalan and Bilgin Özpeynirci 2009) |
| 190 | UTADIS-NM | A sorting method implementing a threshold-based value-driven procedure. It uses indirect preferences from the DMs in the form of incomplete assignments and partial requirements of (non-)monotonicity for each marginal value function. | (Kadziński et al. 2020b) |
| 191 | UTADIS-Regularization | Extension of the UTADIS method with a regularization framework which allows to handle non-monotonic criteria. | (Liu et al. 2019) |
| 192 | UTADIS-ROBUST-REPRESENT | Extension of UTADIS GMS for sorting problems, to derive a unique representative function from the results of the robust ordinal regression. | (Greco et al. 2011b) |
| 193 | UTASTAR | An extension of the UTA method to deal with ordinal criteria. | (Siskos et al. 2016) |
| 194 | Valued-UTADIS | Extension of the UTADIS method by accepting the assignment of reference alternatives to multiple classes with variable credibility degrees. | (Liu et al. 2020b) |
| 195 | VIKOR | This method exploits the difference from an ideal solution and a non-ideal one, together with the weights set by the DM, to rank the alternatives. The difference with the TOPSIS method resides in the approach used to normalize the data and the aggregation function used to derive the ranking. | (Acuña-Soto Claudia 2019) |
| 196 | VIP | An additive value function-based method accepting imprecise information on the criteria performances and preferences. | (Dias and Climaco 2000) |
| 197 | WAM with performances transformation | Weighted additive mean (WAM) based on normalized performances using a data-driven approach. | (Langhans et al. 2014) |
| 198 | WAM without performances transformation | Weighted additive mean (WAM) without any transformation of performances, obtained by the product of raw performances and respective weight. | (Itsubo 2015) |
| 199 | WASPAS | Combination of additive weighted sum and weighted product, with the possibility of varying the degree of contribution from each aggregation function. | (Chakraborty and Zavadskas 2014) |
| 200 | WGM with performances transformation | Weighted geometric mean (WGM) based on normalized performances using a data-driven approach. | (Langhans et al. 2014) |
| 201 | WGM without performances transformation | Weighted geometric mean (WGM) without any transformation of performances, obtained by the product of raw performances and respective weight. | (Itsubo 2015) |
| 202 | WHM with performances transformation | Weighted harmonic mean (WHM) based on normalized performances using a data-driven approach. | (Langhans et al. 2014) |
| 203 | WHM without performances transformation | Weighted harmonic mean (WHM) without any transformation of performances, obtained by the product of raw performances and respective weight. | (Itsubo 2015) |
| 204 | ZAPROS III | A partial ranking method that uses the ordinal information from the criteria performances and the intensities of preference for all neighbouring pairs of performances for all pairs of criteria. | (Larichev 2001) |
| 205 | ZAPROS-LM | A method for partial ranking of alternatives using only the ordinal information on the criteria performances. It uses preference information in the form of pairwise comparisons of alternatives differing with respect to performances on two criteria. | (Larichev and Moshkovich 1997b) |

# Appendix B – Mapping of the MCDA methods, according to the taxonomy, used in the MCDA-MSS

Click here.

# Appendix C – Mapping of the literature case studies, according to the taxonomy, used for the MCDA-MSS test

Click here.